%% file: main.tex
\documentclass[sigconf]{acmart}

\AtBeginDocument{%
  \providecommand\BibTeX{{%
    \normalfont B\kern-0.5em{\scshape i\kern-0.25em b}\kern-0.8em\TeX}}}
\newcommand{\Eqref}[1]{Eq. (\ref{#1})}
\usepackage{breqn}

\copyrightyear{2020} 
\acmYear{2020} 
\setcopyright{acmcopyright}
\acmConference[WSDM '20]{The Thirteenth ACM International Conference on Web Search and Data Mining}{February 3--7, 2020}{Houston, TX, USA}
\acmBooktitle{The Thirteenth ACM International Conference on Web Search and Data Mining (WSDM '20), February 3--7, 2020, Houston, TX, USA}
\acmPrice{15.00}
\acmDOI{10.1145/3336191.3371783}
\acmISBN{978-1-4503-6822-3/20/02}
\begin{document}
\fancyhead{}
\title{Unbiased Recommender Learning from \\
Missing-Not-At-Random Implicit Feedback}

\author{Yuta Saito}
\affiliation{Tokyo Institute of Technology}
\email{saito.y.bj@m.titech.ac.jp}

\author{Suguru Yaginuma}
\affiliation{SMN Corporation}
\email{suguru_yaginuma@so-netmedia.jp}

\author{Yuta Nishino}
\affiliation{SMN Corporation}
\email{yuta_nishino@so-netmedia.jp}

\author{Hayato Sakata}
\affiliation{SMN Corporation}
\email{hayato_sakata@so-netmedia.jp}

\author{Kazuhide Nakata}
\affiliation{Tokyo Institute of Technology}
\email{nakata.k.ac@m.titech.ac.jp}

\renewcommand{\shortauthors}{Saito, et al.}

\begin{abstract}
Recommender systems widely use implicit feedback such as click data because of its general availability. Although the presence of clicks signals the users' preference to some extent, the lack of such clicks does not necessarily indicate a negative response from the users, as it is possible that the users were not exposed to the items (positive-unlabeled problem). This leads to a difficulty in predicting the users' preferences from implicit feedback. Previous studies addressed the positive-unlabeled problem by uniformly upweighting the loss for the positive feedback data or estimating the confidence of each data having relevance information via the EM-algorithm. However, these methods failed to address the missing-not-at-random problem in which popular or frequently recommended items are more likely to be clicked than other items even if a user does not have a considerable interest in them. To overcome these limitations, we first define an ideal loss function to be optimized to realize recommendations that maximize the relevance and propose an unbiased estimator for the ideal loss. Subsequently, we analyze the variance of the proposed unbiased estimator and further propose a clipped estimator that includes the unbiased estimator as a special case. We demonstrate that the clipped estimator is expected to improve the performance of the recommender system, by considering the bias-variance trade-off. We conduct semi-synthetic and real-world experiments and demonstrate that the proposed method largely outperforms the baselines. In particular, the proposed method works better for rare items that are less frequently observed in the training data. The findings indicate that the proposed method can better achieve the objective of recommending items with the highest relevance.
\end{abstract}




\maketitle

\section{Introduction}
Recommender systems are widely used in industries such as personalized advertising and video streaming \cite{2018ImplicitReview,2010LiuClick,bonner2018causal}. These services utilize recommender systems to predict and provide users with items that they may be interested in to improve user experience. To provide such recommendations, predicting the items that users like is an important task \cite{2018ImplicitReview,liang2016factorization,hu2008collaborative}.

 Generally, one can use two types of data for the recommendation. One includes users' ratings on items, and the other corresponds to users' clicks (e.g., purchases, views) \cite{2018ImplicitReview,liang2016modeling}. Rating data is called explicit data, as they represent the preferences explicitly via positive or negative feedback. Typically, collecting sufficient explicit data for recommendation is difficult because it requires users to actively provide ratings \cite{2018ImplicitReview}.
 In contrast, click data, which is called implicit data, are easy to collect because they represent the behavior logs of the users \cite{2010LiuClick}. Any services, even those without the users' ratings, can utilize the data, and many of the actual recommender systems are based on implicit data \cite{2018ImplicitReview,2018Wangsocialexposure:}.
 As in the information retrieval field \cite{wang2018position,joachims2017unbiased,joachims2016counterfactual,liang2016modeling}, one should estimate the relevance of an item to a user from implicit feedback to improve user experience, because one can recommend items that users are genuinely interested in by accurately predicting its relevance.
 However, as the implicit data is not the explicit feedback of the users' preferences, one cannot know whether unclicked feedback is negative feedback or unlabeled positive feedback; this corresponds to the positive-unlabeled problem \cite{2018ImplicitReview,elkan2008learning,bekker2018beyond}.
 
 To predict the relevance, \cite{hu2008collaborative} proposed weighted matrix factorization (WMF), which assigns unclicked items less weight to incorporate the idea that those items correspond to less confidence in prediction than clicked items. However, in some cases, one might be more confident in predicting the relevance of some unclicked items than for the other unclicked ones. Unclicked items that have been recommended several times indicate that they are less relevant for the users. Exposure matrix factorization (ExpoMF) fully utilizes this exposure information \cite{liang2016modeling}. The authors introduced exposure variables and the latent probabilistic model, in which the probability of a click is the product of the probability of the exposure and relevance level. Under this probabilistic model, when a user has been exposed to an item, clicked and unclicked information between the pair can be regarded as relevance information. Thus, ExpoMF upweights the loss of data with high exposure probability because the exposure probability is regarded as the confidence of how much relevance information each data includes. Although ExpoMF tackles the positive-unlabeled problem by introducing the exposure variables, it does not address another critical issue of implicit feedback recommendation, namely the missing-not-at-random (MNAR) problem \cite{yang2018unbiased,schnabel2016recommendations,liu2019spiral}. This is because, by upweighting the loss of data with high exposure probability (mostly popular items), it can lead to poor prediction accuracy for rare items. Thus, ExpoMF suffers from bias and cannot achieve the objective of recommender systems to provide users with items that are relevant to them.
 
 In this study, to establish a method to predict items with the highest relevance rather than the highest click probability, we first define the ideal loss function that should be optimized to maximize the relevance. We theoretically demonstrate that the existing methods (e.g., WMF, ExpoMF) have biases in their loss functions toward the ideal loss. Next, we simultaneously solve the positive-unlabeled problem and the MNAR problem inspired by the estimation technique for causal inference \cite{rubin1974estimating,rosenbaum1983central,imbens2015causal} and derive an unbiased estimator for the ideal loss estimated only from the observable feedback. Further, we analyze the variance of the proposed unbiased estimator and highlight that the variance could be large under recommendation settings. Moreover, we propose a clipped estimator that could achieve a better bias-variance trade-off than that of the unbiased estimator and investigate its theoretical characteristics. Finally, in the experiments, we verify the effectiveness of the proposed method on both semi-synthetic and real-world datasets. 
 
The contributions of this research are as follows.
\begin{itemize}
\item We define an ideal loss function to be optimized to maximize relevance. We propose an unbiased estimator for the loss function of interest for the first time.
\item We perform theoretical analyses for the statistical properties of the proposed estimator and indicate that its variance could be large. We address the variance problem by using a variance reduction estimator.
\item We conduct experiments using semi-synthetic and real-world datasets. Compared to the baselines, the proposed method largely improves the ranking metrics. In addition, it improves the prediction for the less popular items, which comprise a major proportion of the total items.
\end{itemize}

\section{Notation and Problem Formulation}
In this section, we introduce the basic notation and formulate the implicit feedback recommendation.

\subsection{Notation}
Let $u \in \mathcal{U}$ be a user ($|\mathcal{U}| = m$), $i \in \mathcal{I}$ be an item ($ |\mathcal{I}| = n$), and $\mathcal{D} = \mathcal{U} \times \mathcal{I}$ be the set of all user--item pairs. 
$\boldsymbol{Y} \in \{0, 1\}^{m \times n}$ is a click matrix, and each entry $Y_{u,i}$ is a Bernoulli random variable representing a click between the user $u$ and item $i$.
If the feedback of $(u, i)$ is observed, then $Y_{u,i} = 1$; otherwise, $Y_{u, i} = 0$.
In implicit feedback recommendation, $Y_{u,i}=1$ indicates a positive feedback, but
$Y_{u,i} = 0$ is either a negative or unlabeled positive feedback.
To precisely formulate this implicit feedback setting, we introduce two other matrices. $\boldsymbol{R} \in \{0, 1\}^{m \times n}$ is a relevance matrix, and each entry $R_{u, i}$ is a Bernoulli random variable representing the relevance of user $u$ and item $i$.
$R_{u, i} = 1$ means that $u$ and $i$ are relevant; however, $R_{u, i} = 0$ means that $u$ and $i$ are irrelevant. The other matrix is called the exposure matrix, denoted as $\boldsymbol{O} \in \{0, 1\}^{m \times n}$. Each entry of this matrix $O_{u,i}$ is a random variable representing whether user $u$ has been exposed to item $i$. It should be noted that in implicit feedback recommendation, both the relevance and exposure random variables are {\bf unobserved}, and only click random variables are observable.

In this work, we make the following two assumptions for all user--item pairs.
\begin{align}
   Y_{u,i} & = O_{u,i} \cdot R_{u,i} \label{eq1} \\
  P(Y_{u,i} = 1  ) & = \theta_{u,i}  \cdot \gamma_{u,i} \label{eq2}
\end{align}
We define $\theta_{u,i} = P \left( O_{u,i} = 1 \right)$ and $\gamma_{u,i} = P \left( R_{u,i} = 1 \right)$ as the exposure and relevance parameters, respectively and assume that both parameters are non-zero for all user--item pairs.

\Eqref{eq1} assumes that item $i$ is clicked by user $u$ if $i$ has been exposed to $u$ and they are relevant (i.e., $ Y_{u,i} = 1  \Leftrightarrow O_{u,i} = 1  = 1 \; \& \; R_{u,i} = 1 $ ). The position-based model in unbiased learning-to-rank \cite{joachims2017unbiased,wang2018position} makes the same assumption in the click generative process. This assumption precisely represents the implicit feedback setting, in which a click does not always signify relevance.

In contrast, \Eqref{eq2} assumes that the click probability is decomposed into the exposure probability ($ \theta_{u,i} $ ) and relevance level ($\gamma_{u,i}  $)\footnote{This assumption is similar to the unconfoundedness assumption in causal inference \cite{rubin1974estimating,rosenbaum1983central,imbens2015causal} and can also be represented as  $ O \perp R | u, i $.}. Given this assumption, the exposure probability $\theta_{u,i}$ can take different values among user--item pairs, and it can model the MNAR setting in which the click probability and relevance level are not proportional.

\subsection{True Performance Metric}
This section describes the objective of this study. To evaluate the recommendation policy with implicit feedback data, the top-N recommendation metrics such as the mean average precision (MAP) and recall are commonly used \cite{yang2018unbiased}. In general, these metrics can be defined in the following manner \cite{yang2018unbiased}. 
\begin{align}
  \mathcal{R}_{click} \left( \widehat{\boldsymbol{Z}} \right) = \frac{1}{|\mathcal{U}|} \sum_{u \in \mathcal{U}} \,  \sum_{i \in \mathcal{I}} \, \underbrace{P \left( Y_{u,i} = 1 \right)}_{\text{click probability}} \cdot c \left( \widehat{Z}_{u,i} \right) \label{eq3}
\end{align}
where $\widehat{\boldsymbol{Z}} = \{ \widehat{Z}_{u,i} \}_{(u,i) \in \mathcal{D}} $ is the predicted ranking of item $i$ for user $u$, and the function $c(\cdot)$ characterizes a top-N scoring metric. For example, for DCG@K, the function is defined as $c ( \widehat{Z}_{u,i} ) = \mathbb{I} \{ \widehat{Z}_{u,i} \le K \} / \log(\widehat{Z}_{u,i} + 1  )  $. 

The problem is that click ($Y_{u,i}$) does not directly signify relevance ($R_{u,i}$), and thus, the top-N recommendation metrics defined as in \Eqref{eq3} are not appropriate to measure the improvement in the user experience. Therefore, we use the following top-N recommendation metric defined using the relevance level as the performance metric.
\begin{align}
  \mathcal{R}_{rel} \left( \widehat{\boldsymbol{Z}} \right) = \frac{1}{|\mathcal{U}|} \sum_{u \in \mathcal{U}} \, \sum_{i \in \mathcal{I}} \, \underbrace{P \left( R_{u,i} = 1 \right)}_{{\text{relevance level}}} \cdot c \left( \widehat{Z}_{u,i} \right) \label{eq4}
\end{align}

The focus of this study is to optimize the performance metric in \Eqref{eq4}. To achieve this goal, we follow the basic pointwise approach and aim to optimize the following loss function of interest.
\begin{dmath}
 \mathcal{L}_{ideal} \left( \widehat{ \boldsymbol{R}} \right)
   = \frac{1}{ | \mathcal{D} |  } \sum_{(u,i) \in \mathcal{D}} \left[ \gamma_{u,i} \delta^{(1)} \left( \widehat{R}_{u,i} \right) +  \left( 1 - \gamma_{u,i} \right) \delta^{(0)} \left( \widehat{R}_{u,i} \right) \right] \label{eq5}
\end{dmath}
where $\widehat{ \boldsymbol{R}}$ is a prediction matrix, and $ \delta^{(R)}, \; (R \in \{0, 1\}) $ denotes the local loss for user--item pair $(u,i)$. For example, when $\delta^{(R)} (\widehat{R} ) = - ( R \log (\widehat{R} ) + (1 - R) \log (1 - \widehat{R} ) )$, then $ \mathcal{L}_{ideal}( \widehat{ \boldsymbol{R}} ) $ is called the log loss. In the following sections, we denote $ \delta^{(R)} ( \widehat{R}_{u,i} ) $ simply as $\delta^{(R)}_{u,i}$.

A prediction matrix $\widehat{ \boldsymbol{R}}$ minimizing the ideal loss defined using the relevance level in \Eqref{eq5} is expected to lead to the desired values of the top-N recommendation metric in \Eqref{eq4}.

\section{Analysis on Existing Baselines}
In this section, we describe the standard baselines and theoretically analyze the loss functions used in these methods. In particular, we demonstrate that the loss functions are biased against the ideal loss.

\subsection{Weighted Matrix Factorization}
WMF is the most basic latent factor model for implicit feedback recommendation \cite{hu2008collaborative}. The WMF uses a simple heuristic in which all the clicked data are equally upweighted compared with the unclicked data \cite{hu2008collaborative,liang2016modeling}. This model optimizes the following estimator for the ideal loss:
\begin{align}
  \widehat{\mathcal{L}}_{WMF} \left( \widehat{ \boldsymbol{R}} \right)
  = \frac{1}{ | \mathcal{D} | } \sum_{(u,i) \in \mathcal{D}} \left[ c Y_{u,i} \delta^{(1)}_{u,i} + \left( 1 - Y_{u,i} \right) \delta^{(0)}_{u,i} \right] \label{eq6}
\end{align}
where $c \ge 1$ is a hyperparameter representing the confidence of the clicked data relative to that of the unclicked data. When no side information is available, $c$ is uniform among all the clicked data. In the following proposition, we demonstrate that the estimator used in the WMF is biased.
\begin{proposition}(Bias of WMF estimator)
    The bias of the estimator used in WMF is represented as follows.
    \begin{align*}
         & \left| \mathbb{E}\left[ \widehat{ \mathcal{L}}_{WMF} \left( \widehat{ \boldsymbol{R}} \right) \right] - \mathcal{L}_{ideal} \left( \widehat{ \boldsymbol{R}} \right)  \right| \\
         & = 
         \frac{1}{|\mathcal{D}|} \left| \sum_{(u,i) \in \mathcal{D}} \left[ (c \theta_{u,i} - 1) \gamma_{u,i} \delta^{(1)} + \gamma_{u,i} (1 - \theta_{u,i}) \delta_{u,i}^{(0)} \right] \right| 
    \end{align*}
  \begin{proof}
    \begin{align}
      \mathbb{E}\left[ \widehat{ \mathcal{L}}_{WMF} \left( \widehat{ \boldsymbol{R}} \right) \right]
      & = \mathbb{E}\left[ \frac{1}{ | \mathcal{D} | } \sum_{(u,i) \in \mathcal{D}} \left[ c Y_{u,i} \delta^{(1)}_{u,i} + \left( 1 - Y_{u,i} \right) \delta^{(0)}_{u,i} \right] \right] \notag \\
      & =  \frac{1}{ | \mathcal{D} | } \sum_{(u,i) \in \mathcal{D}} \left[ c \mathbb{E}\left[ Y_{u,i} \right] \delta^{(1)}_{u,i} + \left( 1 - \mathbb{E}\left[ Y_{u,i} \right] \right) \delta^{(0)}_{u,i} \right]  \notag \\
      & = \frac{1}{ | \mathcal{D} | } \sum_{(u,i) \in \mathcal{D}} \left[ c \theta_{u,i} \gamma_{u,i} \delta^{(1)}_{u,i} + \left( 1 - \theta_{u,i} \gamma_{u,i} \right) \delta^{(0)}_{u,i} \right]  \notag
    \end{align}
    Thus,
    \begin{dmath*}
         \mathbb{E}\left[ \widehat{ \mathcal{L}}_{WMF} \left( \widehat{ \boldsymbol{R}} \right) \right] - \mathcal{L}_{ideal} \left( \widehat{ \boldsymbol{R}} \right)
         = \frac{1}{ | \mathcal{D} | } \sum_{(u,i) \in \mathcal{D}} \left[ c \theta_{u,i} \gamma_{u,i} \delta^{(1)}_{u,i} + \left( 1 - \theta_{u,i} \gamma_{u,i} \right) \delta^{(0)}_{u,i} \right]
         - \frac{1}{ | \mathcal{D} | } \sum_{(u,i) \in \mathcal{D}} \left[ \gamma_{u,i} \delta^{(1)}_{u,i} + \left( 1 - \gamma_{u,i} \right) \delta^{(0)}_{u,i}  \right]
         = \frac{1}{|\mathcal{D}|} \sum_{(u,i) \in \mathcal{D}} \left[ (c \theta_{u,i} - 1) \gamma_{u,i} \delta^{(1)} + \gamma_{u,i} (1 - \theta_{u,i}) \delta_{u,i}^{(0)} \right] 
    \end{dmath*}
  \end{proof}
\end{proposition}

For $ \widehat{ \mathcal{L}}_{WMF} \left( \widehat{ \boldsymbol{R}} \right)$ to be theoretically unbiased, $c\theta_{u,i} - 1 = 0 \Rightarrow \theta_{u,i} = 1 / c$, and $1 - \theta_{u,i} = 0 \Rightarrow \theta_{u,i} = 1$ need to be satisfied for all pairs from the last equation. However,  $\theta_{u,i}$ can take different values among user--item pairs in our setting; thus, these conditions are not always satisfied, and the loss function of WMF does not satisfy the unbiasedness. This is because WMF does not address the positive-unlabeled problem, in which $Y = 0$ does not always signify $R = 0$. Thus, WMF is unsuitable for optimizing our metric of interest defined in \Eqref{eq4}.

\subsection{Exposure Matrix Factorization}
In contrast to the WMF, the ExpoMF is a prediction method considering the exposure matrix ($\boldsymbol{O}$), and it is based on the following latent probabilistic model \cite{liang2016modeling}:
\begin{align*}
  \boldsymbol{U} & \; \sim \; \mathcal{N}\left(\mathbf{0}, \lambda_{U}^{-1} I_{K}\right), \; \boldsymbol{V}  \; \sim \; \mathcal{N}\left(\mathbf{0}, \lambda_{V}^{-1} I_{K}\right)  \\
  O_{u, i} & \; \sim \; Bern \left(\mu_{i}\right),  \;
  Y_{u, i} \, | \, O_{u, i}=1  \; \sim \; \mathcal{N}\left(U_{u}^{\top} V_{i}, \lambda_{y}^{-1}\right) 
\end{align*}
where $\lambda_{U}$, $\lambda_{V}$, and $\lambda_{y}$ are the hyperparameters denoting the inverse variance of the prior distributions, and $Bern(\cdot)$ is the Bernoulli distribution. Following the probabilistic model defined above, $P( Y_{u,i} = 0 \, | \, O_{u,i} = 0 ) = 1 $, which is consistent with our assumptions.

The log-likelihood to derive the parameters (i.e., $\mu_i$, $U_u$, and $V_i$) can be written as\footnote{Equation (2) of \cite{liang2016modeling}. The third term is always $0$ and is thus omitted here.}
\begin{align}
  & \log \left(P\left(o_{u, i}, y_{u, i} \, | \, \mu_{u, i}, U_{u}, V_{i}, \lambda_{y}^{-1}\right)\right) \notag \\
  &  \quad =\log Bern \left(o_{u, i} \, | \, \mu_{u, i}\right)
  + \underbrace{o_{u, i} \cdot \log \mathcal{N} \left(y_{u, i} \, | \, U_{u}^{\top} V_{i}, \lambda_{y}^{-1}\right)}_{(a)} \notag \label{eq7} \\
\end{align}
In \Eqref{eq7}, the loss to derive the user and item matrices ($\boldsymbol{U}, \boldsymbol{V} $) is $(a)$, and it can be defined in the following manner:
\[
  (a) = \begin{cases}
    \delta^{(1)} \left( U_{u}^{\top} V_{i} \right) & (y_{u,i}=1 , \, o_{u,i} = 1)  \\
    \delta^{(0)} \left( U_{u}^{\top} V_{i} \right) & (y_{u,i}=0 , \, o_{u,i} = 1) \\
    0 & \text{otherwise } (o_{u,i} = 0)
  \end{cases}
\]
Therefore, the loss function of the ExpoMF is designed to consider the local loss of user--item pairs when the item has been exposed to the user (i.e., $o_{u,i}=1$). This is because if an item has been exposed, the click variable is considered to represent the relevance information (i.e., $O_{u,i} = 1  \Rightarrow Y_{u,i} = R_{u,i}$).

However, the realizations of exposure variables $\{ o_{u,i}\}$ are unobserved. Therefore, ExpoMF uses an EM-like iterative algorithm to derive the user--item matrices. In the E-step, the posterior exposure probability is estimated, and in the M-step, the model parameters are updated to optimize the log-likelihood\footnote{The detailed learning procedure of ExpoMF is described in Section 3.3 of \cite{liang2016modeling}.}. Given the true posterior exposure probabilities, the M-step of ExpoMF is interpreted to minimize the following weighted loss function.
 \begin{align}
  \widehat{\mathcal{L}}_{ExpoMF} \left( \widehat{ \boldsymbol{R}} \right)
  & = \frac{1}{ | \mathcal{D} | } \sum_{(u,i) \in \mathcal{D}} \theta^{\prime}_{u,i} \left[   Y_{u,i} \delta^{(1)}_{u,i} + \left( 1 - Y_{u,i} \right) \delta^{(0)}_{u,i}  \right]  \label{eq8}
 \end{align}
 where $\theta^{\prime}_{u,i} = \mathbb{E} \left[ O_{u,i} \, | \, Y_{u,i} \right] $ is the posterior exposure probability. For example, $\mathbb{E} \left[ O_{u,i} \, | \, Y_{u,i} = 1 \right] = 1 $ because $Y_{u,i} = 1 \, \Rightarrow O_{u,i} = 1$. This posterior probability represents the confidence of the amount of relevance information contained in click indicator $Y_{u,i}$.
 
 ExpoMF utilizes the posterior probability of the exposure to reweight data and improves the WMF. However, the following proposition suggests that the loss function in \Eqref{eq8} optimized in the M-step of the ExpoMF is also biased against the ideal loss.
 
 \begin{proposition}(Bias of ExpoMF) 
    When the posterior exposure probabilities are given, then, the bias of the estimator used in ExpoMF is represented as follows.
     \begin{align*}
         & \left| \mathbb{E}\left[ \widehat{ \mathcal{L}}_{ExpoMF} \left( \widehat{ \boldsymbol{R}} \right) \right] - \mathcal{L}_{ideal} \left( \widehat{ \boldsymbol{R}} \right)  \right| \\
         & = 
         \frac{1}{ | \mathcal{D} | } \Bigl| \sum_{(u,i) \in \mathcal{D}}  \gamma_{u,i} ( \theta^{\prime}_{u,i} \theta_{u,i} - 1 )\delta^{(1)}_{u,i} \\
         & \qquad \qquad \qquad  + \left\{ \theta^{\prime}_{u,i} - 1 - \gamma_{u,i} (\theta_{u,i} \theta^{\prime}_{u,i} - 1 )  \right\} \delta^{(0)}_{u,i}  \Bigr|
    \end{align*}
   \begin{proof}
     \begin{dmath*}
       \mathbb{E}\left[ \widehat{ \mathcal{L}}_{ExpoMF} \left( \widehat{ \boldsymbol{R}} \right) \right] \notag \\
       = \mathbb{E}\left[ \frac{1}{ | \mathcal{D} | } \sum_{(u,i) \in \mathcal{D}}  \theta^{\prime}_{u,i} \left[  Y_{u,i} \delta^{(1)}_{u,i} + \left( 1 - Y_{u,i} \right) \delta^{(0)}_{u,i}  \right] \right] 
       =  \frac{1}{ | \mathcal{D} | } \sum_{(u,i) \in \mathcal{D}} \theta^{\prime}_{u,i} \left[ \mathbb{E}\left[ Y_{u,i} \right] \delta^{(1)}_{u,i} +   \left( 1 -  \mathbb{E}\left[ Y_{u,i} \right] \right) \delta^{(0)}_{u,i}  \right]
       =  \frac{1}{ | \mathcal{D} | } \sum_{(u,i) \in \mathcal{D}} \left[ \theta^{\prime}_{u,i} \theta_{u,i} \gamma_{u,i} \delta^{(1)}_{u,i} +  \theta^{\prime}_{u,i} \left( 1 -  \theta_{u,i} \gamma_{u,i} \right) \delta^{(0)}_{u,i}  \right] 
     \end{dmath*}
     Thus we obtain:
     \begin{dmath*}
         \mathbb{E}\left[ \widehat{ \mathcal{L}}_{ExpoMF} \left( \widehat{ \boldsymbol{R}} \right) \right] - \mathcal{L}_{ideal} \left( \widehat{ \boldsymbol{R}} \right)  
           = \frac{1}{ | \mathcal{D} | } \sum_{(u,i) \in \mathcal{D}} \left[ \theta^{\prime}_{u,i} \theta_{u,i} \gamma_{u,i} \delta^{(1)}_{u,i} +  \theta^{\prime}_{u,i} \left( 1 -  \theta_{u,i} \gamma_{u,i} \right) \delta^{(0)}_{u,i} \right]
           - \frac{1}{ | \mathcal{D} | } \sum_{(u,i) \in \mathcal{D}} \left[ \gamma_{u,i} \delta^{(1)}_{u,i} + \left( 1 - \gamma_{u,i} \right) \delta^{(0)}_{u,i}  \right]
           = \frac{1}{ | \mathcal{D} | } \sum_{(u,i) \in \mathcal{D}} \left[ \gamma_{u,i} ( \theta^{\prime}_{u,i} \theta_{u,i} - 1 )\delta^{(1)}_{u,i} 
           + \left\{ \theta^{\prime}_{u,i} - 1 - \gamma_{u,i} (\theta_{u,i} \theta^{\prime}_{u,i} - 1 )  \right\} \delta^{(0)}_{u,i} \right] 
     \end{dmath*}
   \end{proof}
 \end{proposition}
 For $ \widehat{ \mathcal{L}}_{ExpoMF} \left( \widehat{ \boldsymbol{R}} \right)$ to be theoretically unbiased, $\theta_{u,i} \theta^{\prime}_{u,i} - 1 = 0 \Rightarrow \theta_{u,i} \theta^{\prime}_{u,i} = 1$, and $ \theta^{\prime}_{u,i}  - 1 = 0 \Rightarrow \theta^{\prime}_{u,i} = 1$ need to be satisfied for all pairs from the last equation. However, $\theta_{u,i}$ and $\theta^{\prime}_{u,j}$ can take different values among user--item pairs in our setting, and thus, these conditions are not always satisfied, and the loss function of the ExpoMF does not satisfy the unbiasedness. This is because the ExpoMF upweights the local loss of data that is frequently observed in the training data (i.e., data having high exposure probability). This upweighting leads to the poor prediction accuracy for data having low exposure probability such as tail items, and thus, it fails to achieve the goal of recommender systems. Therefore, dealing with the MNAR problem as well as the unlabeled nature of implicit feedback is essential to derive a desirable estimator.

\section{Proposed Method}
In this section, we propose an unbiased estimator for the ideal loss to overcome the limitations described in the previous section. The proposed unbiased estimator is an extension of the Inverse Propensity Score (IPS) in causal inference \cite{rosenbaum1983central,rubin1974estimating,imbens2015causal} and an estimator in positive-unlabeled learning \cite{bekker2018beyond,li2005learning}.
In our theoretical analysis, we prove that our estimator is valid in the implicit recommendation setting. Moreover, we analyze the variance of the unbiased estimator and indicate that it can suffer from a high variance. Finally, we provide and analyze a technique to address the variance problem.

\subsection{Proposed Estimator}
First, we define the propensity score to deal with the MNAR problem of implicit feedback as follows.
\begin{definition}(Propensity Score) The propensity score of user--item pair $(u,i)$ is
  \begin{align}
    \theta_{u,i} = P ( O_{u,i} = 1 ) = P  \left( Y_{u,i} = 1 \, | \, R_{u,i} = 1  \right) \notag
  \end{align}
\end{definition}

Next, our proposed estimator is defined using the propensity score.
\begin{definition}(Unbiased Estimator) When the propensity scores are given, the unbiased estimator is defined as
  \begin{align}
      \widehat{\mathcal{L}}_{unbiased} \left( \widehat{ \boldsymbol{R}} \right)  = \frac{1}{| \mathcal{D} |} \sum_{(u,i) \in \mathcal{D}} \left[  \frac{Y_{u,i}}{\theta_{u,i}} \delta^{(1)}_{u,i}  + \left(1 - \frac{Y_{u,i}}{\theta_{u,i}} \right) \delta^{(0)}_{u,i} \right] \label{eq9}
  \end{align}
\end{definition}
Note that this unbiased estimator is not the standard form of the IPS estimator in the explicit feedback setting \cite{schnabel2016recommendations} because one cannot use the exposure indicator in the implicit feedback setting. The proposed estimator can also be represented in the following form.
\begin{align*}
      \frac{1}{| \mathcal{D} |} \sum_{(u,i) \in \mathcal{D}} \left[  Y_{u,i} \left( \frac{1}{\theta_{u,i}} \delta^{(1)}_{u,i} + \left(1 - \frac{1}{\theta_{u,i}} \delta^{(0)}_{u,i}  \right) \right)  + ( 1 - Y_{u,i}) \delta^{(0)}_{u,i} \right]
  \end{align*}

Thus, the unbiased estimator applies both positive loss ($\delta^{(1)}$) and negative loss ($\delta^{(0)}$) for the clicked data ($Y_{u,i} = 1$). In the following proposition, we show that our unbiased estimator is unbiased against the ideal loss.
\begin{proposition}
    The unbiased estimator defined in \Eqref{eq9} is unbiased against the ideal loss in \Eqref{eq5}.
  \begin{align}
    \mathbb{E} \left[ \widehat{\mathcal{L}}_{unbiased} \left( \widehat{ \boldsymbol{R}} \right)  \right] = \mathcal{L}_{ideal} \left( \widehat{ \boldsymbol{R}} \right) \notag 
  \end{align}
  \begin{proof}
    \begin{align*}
       & \mathbb{E} \left[ \widehat{\mathcal{L}}_{unbiased} \left( \widehat{ \boldsymbol{R}} \right)  \right]  \\
       & = \mathbb{E} \left[ \frac{1}{| \mathcal{D} |} \sum_{(u,i) \in \mathcal{D}} \left[  \frac{Y_{u,i}}{\theta_{u,i}} \delta^{(1)}_{u,i} + \left(1 - \frac{Y_{u,i}}{\theta_{u,i}} \right) \delta^{(0)}_{u,i} \right] \right] \\
       & = \frac{1}{| \mathcal{D} |} \sum_{(u,i) \in \mathcal{D}} \left[  \frac{ \mathbb{E} [ Y_{u,i} ]}{\theta_{u,i}} \delta^{(1)}_{u,i} + \left(1 - \frac{\mathbb{E} [ Y_{u,i}]}{\theta_{u,i}} \right) \delta^{(0)}_{u,i} \right] \\
      & = \frac{1}{ | \mathcal{D} |  } \sum_{(u,i) \in \mathcal{D}} \left[ \gamma_{u,i} \delta^{(1)}_{u,i} +  \left( 1 - \gamma_{u,i} \right) \delta^{(0)}_{u,i} \right] = \mathcal{L}_{ideal}  \left( \widehat{ \boldsymbol{R}} \right) 
    \end{align*}
  \end{proof}
\end{proposition}

Proposition 4.3 validates that de-biasing using propensity reweighting is valid in MNAR implicit recommendation. However, propensity-based estimators often suffer from a high variance \cite{saito2019doubly,chen2019top,swaminathan2015self}. Moreover, variance analysis was not conducted in a previous study on positive-unlabeled learning \cite{bekker2018beyond,li2005learning}. Thus, in the following theorem, we provide the variance of the unbiased estimator.

\begin{theorem}(Variance of the unbiased estimator) Given sets of independent random variables $ \{ Y_{u,i}\}, \{ O_{u,i}\}$, and $\{ R_{u,i}\} $, propensity scores $\{ \theta_{u,i} \}$, and predicted matrix $\widehat{ \boldsymbol{R}}$, the variance of the unbiased estimator is
  \begin{align}
    \mathbb{V} \left( \widehat{ \mathcal{L}}_{unbiased} \left( \widehat{ \boldsymbol{R}} \right)  \right)
    = \frac{1}{ | \mathcal{D} |^2 } \sum_{(u,i) \in \mathcal{D}}  \gamma_{u,i} \left( \frac{1}{\theta_{u,i}} - \gamma_{u,i} \right) \left(\delta^{(1)}_{u,i} - \delta^{(0)}_{u,i}  \right)^2 \notag 
  \end{align}
  \begin{proof}
    First, we define $X_{u,i} = \frac{Y_{u,i}}{\theta_{u,i}} \delta^{(1)}_{u,i} + \left(1 - \frac{Y_{u,i}}{\theta_{u,i}} \right) \delta^{(0)}_{u,i} $. 
    Subsequently, $ \mathbb{V} \left( X_{u,i} \right) $ can be written as
    \begin{align}
      \mathbb{V} \left( X_{u,i} \right)  = \underbrace{\mathbb{E} \left[ ( X_{u,i} )^2 \right]}_{(b)} -
      \underbrace{\left( \mathbb{E} \left[ X_{u,i} \right] \right)^2}_{(c)} \notag
    \end{align}
     By Proposition 4.3, 
     \begin{dmath*}
        (c) = ( \gamma_{u,i}\delta^{(1)}_{u,i} + (1 - \gamma_{u,i})\delta^{(0)}_{u,i} )^2 = \gamma_{u,i}^2 ( \delta_{u,i}^{(1)} )^2 + 2  \gamma_{u,i} (1 - \gamma_{u,i}) \delta_{u,i}^{(1)} \delta_{u,i}^{(0)} + (1 - \gamma_{u,i})^2 ( \delta_{u,i}^{(0)} )^2
     \end{dmath*}
     Then,
    \begin{dmath*}
      X_{u,i}^2 
      = \frac{Y_{u,i}}{\theta_{u,i}^2}( \delta_{u,i}^{(1)} )^2 + 2 \frac{Y_{u,i}}{\theta_{u,i}} \left(1 - \frac{Y_{u,i}}{\theta_{u,i}} \right) \delta^{(1)}_{u,i} \delta^{(0)}_{u,i} +  \left(1 - \frac{Y_{u,i}}{\theta_{u,i}} \right)^2 ( \delta_{u,i}^{(1)} )^2
      = \frac{Y_{u,i}}{\theta_{u,i}^2}( \delta_{u,i}^{(1)} )^2 + 2 \left(\frac{Y_{u,i}}{\theta_{u,i}}  -  \frac{Y_{u,i}}{\theta_{u,i}^2} \right) \delta^{(1)}_{u,i} \delta^{(0)}_{u,i} + \left(1 - 2\frac{Y_{u,i}}{\theta_{u,i}} + \frac{Y_{u,i}}{\theta_{u,i}^2} \right) ( \delta_{u,i}^{(0)} )^2
    \end{dmath*}
    where $Y_{u,i}^2 = Y_{u,i} $ and $ (1 - Y_{u,i})^2 = (1 - Y_{u,i})$. Next, $(b)$ is calculated as
    \begin{align*}
      (b) 
       = \frac{\gamma_{u,i}}{\theta_{u,i}}( \delta_{u,i}^{(1)} )^2 + 2 \left(\gamma  -  \frac{\gamma}{\theta_{u,i}} \right) \delta^{(1)}_{u,i} \delta^{(0)}_{u,i} + \left(1 - 2\gamma + \frac{\gamma}{\theta_{u,i}} \right) ( \delta_{u,i}^{(0)} )^2
    \end{align*}
    Therefore,
    \begin{align*}
      \mathbb{V} \left( X_{u,i} \right)
       = (b) - (c)  = \gamma_{u,i} \left( \frac{1}{\theta_{u,i}} - \gamma_{u,i} \right) \left(\delta_{u,i}^{(1)} - \delta_{u,i}^{(0)} \right)^2
    \end{align*}
    From the assumptions, $\{  X_{u,i} \}$ is a set of independent random variables. Thus,
    \begin{align}
      \mathbb{V} \left( \widehat{ \mathcal{L}}_{unbiased} \left( \widehat{ \boldsymbol{R}} \right)  \right)
      & = \frac{1}{ | \mathcal{D} |^2 } \sum_{(u,i) \in \mathcal{D}} \mathbb{V} \left( X_{u,i} \right) \notag \\
      & = \frac{1}{ | \mathcal{D} |^2 } \sum_{(u,i) \in \mathcal{D}}  \gamma_{u,i} \left( \frac{1}{\theta_{u,i}} - \gamma_{u,i} \right) \left(\delta^{(1)}_{u,i} - \delta^{(0)}_{u,i}  \right)^2 \notag
    \end{align}
  \end{proof}
\end{theorem}

As shown in Theorem 4.4, the variance of the unbiased estimator depends on the inverse of the propensity score. The propensity score is defined as the exposure probability, and thus it can be small, especially for tail items \cite{yang2018unbiased}. Therefore, the variance of the proposed estimator can be large in the implicit recommendation setting.

\subsection{Variance Reduction Technique}
In the theoretical analysis, we demonstrated that the proposed unbiased estimator is unbiased against the ideal loss. However, Theorem 4.4 suggests that the unbiased estimator can suffer from high variance. In this subsection, we apply the propensity clipping technique \cite{chen2019top,gilotte2018offline,su2019cab} to our estimator and analyze the bias-variance trade-off of an estimator with clipping.

First, we introduce the propensity score with the clipping technique.

\begin{definition}(Clipped Propensity Score) $M$ is a positive constant that takes values in the interval $[0,1]$. Then, the clipped propensity score is defined as:
  \begin{align}
    \bar{\theta}_{u,i} = \max \{ \theta_{u,i}, M \} 
  \end{align}
\end{definition}

The clipped propensity score clips a small value of $\theta$ by $M$. We further define the clipped estimator for the ideal loss by using the clipped propensity score.
\begin{definition}(Clipped Estimator) The clipped estimator for the ideal loss is defined as
  \begin{align}
     \widehat{\mathcal{L}}_{clipped} \left( \widehat{ \boldsymbol{R}} \right)  = \frac{1}{| \mathcal{D} |} \sum_{(u,i) \in \mathcal{D}} \left[  \frac{Y_{u,i}}{\bar{\theta}_{u,i}} \delta^{(1)}_{u,i}  + \left(1 - \frac{Y_{u,i}}{\bar{\theta}_{u,i}} \right) \delta^{(0)}_{u,i} \right]  \label{eq11}
  \end{align}
\end{definition}

As $M \rightarrow 1$, the clipped estimator approaches the naive estimator (estimator of WMF with $c=1$); in contrast, as $M \rightarrow 0$, it approaches the unbiased estimator. Thus, the clipped estimator is a general form of the two estimators. By definition, the inverse of the clipped propensity score is always smaller than that of the propensity score. Thus, using the clipped estimator reduces the effect of the variance problem of the unbiased estimator. However, it introduces some bias because the clipped propensity score is not always equal to the true propensity. We provide the bias and variance of the clipped estimator as follows.

\begin{proposition}(Bias of the clipped estimator) Given a constant $M \in [0,1]$, the bias of the clipped estimator is
  \begin{align}
    & \left| \mathbb{E}\left[ \widehat{ \mathcal{L}}_{clipped} \left( \widehat{ \boldsymbol{R}} \right) \right] - \mathcal{L}_{ideal} \left( \widehat{ \boldsymbol{R}} \right) \right| \notag \\
    & = \left| \frac{1}{|\mathcal{D}|} \sum_{(u,i) \in \mathcal{D}}  \mathbb{I} \{ \theta_{u,i} \le M \} \gamma_{u,i} \left( \frac{\theta_{u,i}}{M} - 1 \right) \left( \delta_{u,i}^{(1)} - \delta_{u,i}^{(0)} \right) \right| \notag
  \end{align}
  \begin{proof}
    First, we define $X_{u,i} = \frac{Y_{u,i}}{\theta_{u,i}} \delta^{(1)}_{u,i} + \left(1 - \frac{Y_{u,i}}{\theta_{u,i}} \right) \delta^{(0)}_{u,i} $, then,
    \begin{align*}
        & \mathbb{E}\left[ \widehat{ \mathcal{L}}_{clipped} \left( \widehat{ \boldsymbol{R}} \right) \right] \\
        & =   \frac{1}{|\mathcal{D}|} \sum_{(u,i) \in \mathcal{D}}  \mathbb{I} \{ \theta_{u,i} > M \} \mathbb{E} [ X_{u,i} ]  \\
        & + \frac{1}{| \mathcal{D} |} \sum_{(u,i) \in \mathcal{D}} \mathbb{I} \{ \theta_{u,i} \le M \} \underbrace{ \mathbb{E} \left[ \frac{ Y_{u,i}}{M}\delta^{(1)}_{u,i}  + \left(1 - \frac{ Y_{u,i}}{M} \right) \delta^{(0)}_{u,i} \right]}_{(d)}
    \end{align*}
    By using $ \mathbb{E} [ Y_{u,i} ] = \theta_{u,i} \gamma_{u,i} $, $(d)$ is calculated as
    \begin{align*}
        (d) =  \frac{\theta_{u,i} \gamma_{u,i}  }{M} \delta^{(1)}_{u,i}  + \left(1 -\frac{\theta_{u,i} \gamma_{u,i} }{M}  \right) \delta_{u,i}^{(0)} 
    \end{align*}
    By Proposition 4.3, $\mathbb{E}\left[  X_{u,i}  \right] = \gamma_{u,i} \delta^{(1)}_{u,i} + (1 - \gamma_{u,i}) \delta^{(0)}_{u,i} $. Then,
    the ideal loss can also be represented as 
    \begin{align*}
        \mathcal{L}_{ideal}  \left( \widehat{ \boldsymbol{R}} \right)
        =  \frac{1}{|\mathcal{D}|} \sum_{(u,i) \in \mathcal{D}}  \mathbb{I} \{ \theta_{u,i} > M \} \mathbb{E} [ X_{u,i} ] + \mathbb{I} \{ \theta_{u,i} \le M \} \mathbb{E} [ X_{u,i} ]
    \end{align*}
    Therefore, 
    \begin{align*}
         & \mathbb{E}\left[ \widehat{ \mathcal{L}}_{clipped} \left( \widehat{ \boldsymbol{R}} \right) \right] - \mathcal{L}_{ideal} \left( \widehat{ \boldsymbol{R}} \right) \\
        & =  \frac{1}{|\mathcal{D}|} \sum_{(u,i) \in \mathcal{D}}  \mathbb{I} \{ \theta_{u,i} \le M \} \left( (d) -  \mathbb{E} [ X_{u,i} ] \right)  \\
        & = \frac{1}{|\mathcal{D}|} \sum_{(u,i) \in \mathcal{D}} \left[ \mathbb{I} \{ \theta_{u,i} \le M \} \gamma_{u,i} \left( \frac{\theta_{u,i}}{M} - 1 \right) \left( \delta_{u,i}^{(1)} - \delta_{u,i}^{(0)} \right)  \right] 
    \end{align*}
  \end{proof}
\end{proposition}

\begin{corollary}(Variance of the clipped estimator) Under the same condition as in Theorem 4.4, the variance of the clipped estimator is
  \begin{align*}
     \mathbb{V} \left( \widehat{ \mathcal{L}}_{clipped} \left( \widehat{ \boldsymbol{R}} \right)  \right) 
    & = \frac{1}{|\mathcal{D}|^2} \sum_{(u,i) \in \mathcal{D}} \gamma_{u,i} \left( \frac{1}{\bar{\theta}_{u,i}} - \gamma_{u,i} \right) \left(\delta^{(1)}_{u,i} - \delta^{(0)}_{u,i}  \right)^2 \\
    & \le \mathbb{V} \left( \widehat{ \mathcal{L}}_{unbiased} \left( \widehat{ \boldsymbol{R}} \right)  \right) \; \because \frac{1}{\bar{\theta}_{u,i}} \le \frac{1}{\theta_{u,i}}
  \end{align*}
  \begin{proof}
    Propensity scores are not random variables. Thus, we obtain the variance by replacing $\theta_{u,i}$ in Theorem 4.4 by $\bar{\theta}_{u,i}$.
  \end{proof}
\end{corollary}

As shown above, the clipped estimator always reduces the variance of the unbiased estimator but introduces some bias depending on the value of $M$.
In the experimental part, we tuned the clipping constant $M$ as a hyper-parameter using a validation set to address the variance problem.

\section{Semi-Synthetic Experiment}
In this section, we conduct an experiment with semi-synthetic datasets\footnote{The code is available at https://github.com/usaito/unbiased-implicit-rec.} and investigate the following research questions (RQs).

\begin{itemize}
    \item [RQ1.] How does the level of exposure bias affect the performance of the MF-Naive and MF-Unbiased models?
    \item [RQ2.] Does optimizing the ideal pointwise loss in \Eqref{eq5} yield a better value of the ranking metric in \Eqref{eq4}?
\end{itemize}

\subsection{Experimental Setup}

\subsubsection{Dataset}
We used the MovieLens (ML) 100K dataset\footnote{http://grouplens.org/datasets/movielens/}. This dataset contains five-star movie ratings collected from a movie recommendation service, and the ratings are MNAR. To facilitate ground-truth evaluation against a fully known relevance and exposure parameters, we created these parameters as follows.

\begin{itemize}
    \item [1.] Using rating-based matrix factorization \cite{koren2009matrix}, we found an approximation of the true ratings as
    \begin{align*}
        R_{u,i} \approx \widehat{R}_{u,i}, \, \forall (u,i) \in \mathcal{D}
    \end{align*}
    where $\widehat{R}_{u,i} \in [1, 5]$.
    \item [2.] Using logistic matrix factorization \cite{johnson2014logistic}, we found an approximation of the true observations as
    \begin{align*}
        O_{u,i} \approx \widehat{O}_{u,i}, \, \forall (u,i) \in \mathcal{D}
    \end{align*}
    where $O_{u,i}$ is a binary variable representing whether the rating of $(u,i)$ is observed. If the rating of $(u,i)$ is observed, $O_{u,i}=1$; otherwise, $O_{u,i}=0$. Thus, $\widehat{O}_{u,i} \in (0, 1)$ is the estimated probability of observing the rating of $(u,i)$.
    \item [3.] We generate the ground-truth relevance and exposure parameters as follows.
    \begin{align*}
        \gamma_{u,i} = \sigma(\widehat{R}_{u,i}  - \epsilon ), \;
        \theta_{u,i} = (\widehat{O}_{u,i})^{p}, \, \forall (u,i) \in \mathcal{D}
    \end{align*}
    where $\sigma(\cdot)$ is the sigmoid function, $\epsilon$ controls the overall relevance level, and $p$ controls the skewness of the distribution of the exposure parameter. When a large value of $p$ is used, then a huge exposure bias is introduced. In the experiment, we set $\epsilon=5$ and $p = 0.5, 1, 2, 3, 4$.
    \item[4. ] Following the probabilistic model described in \Eqref{eq1} and \Eqref{eq2}, we generated click variables as follows.
        \begin{align*}
            O_{u,i} & \sim Bern (\theta_{u,i} ), \, R_{u,i} \sim Bern (\gamma_{u,i} ) \\
            Y_{u,i} & =  O_{u,i} \cdot R_{u,i}, \, \forall (u,i) \in \mathcal{D}
        \end{align*}
    where $Bern(\cdot)$ is the Bernoulli distribution.
\end{itemize}

Note that one can evaluate the ground-truth performance of the methods with the true relevance and exposure parameters by using semi-synthetic datasets.

\begin{figure*}[ht]
    \centering
    \begin{center}
        \begin{tabular}{c}
          \begin{minipage}{0.25\hsize}
                \begin{center}
                    \includegraphics[clip, width=4.5cm]{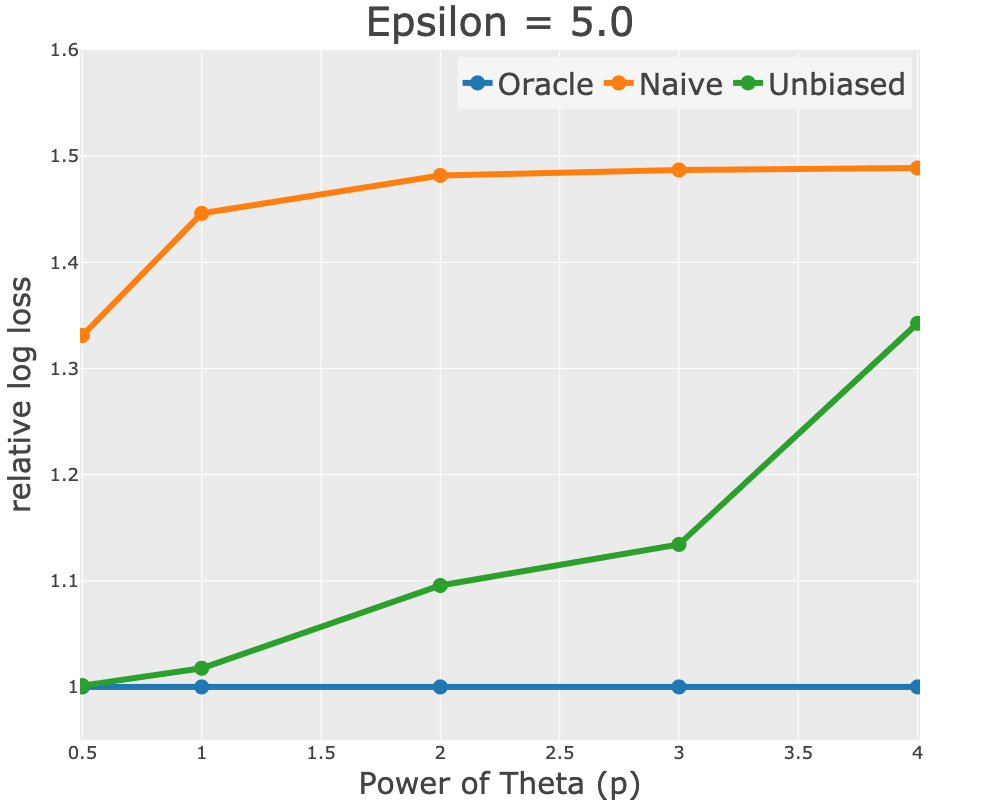}
                \end{center}
            \end{minipage}
        
            \begin{minipage}{0.25\hsize}
                \begin{center}
                    \includegraphics[clip, width=4.5cm]{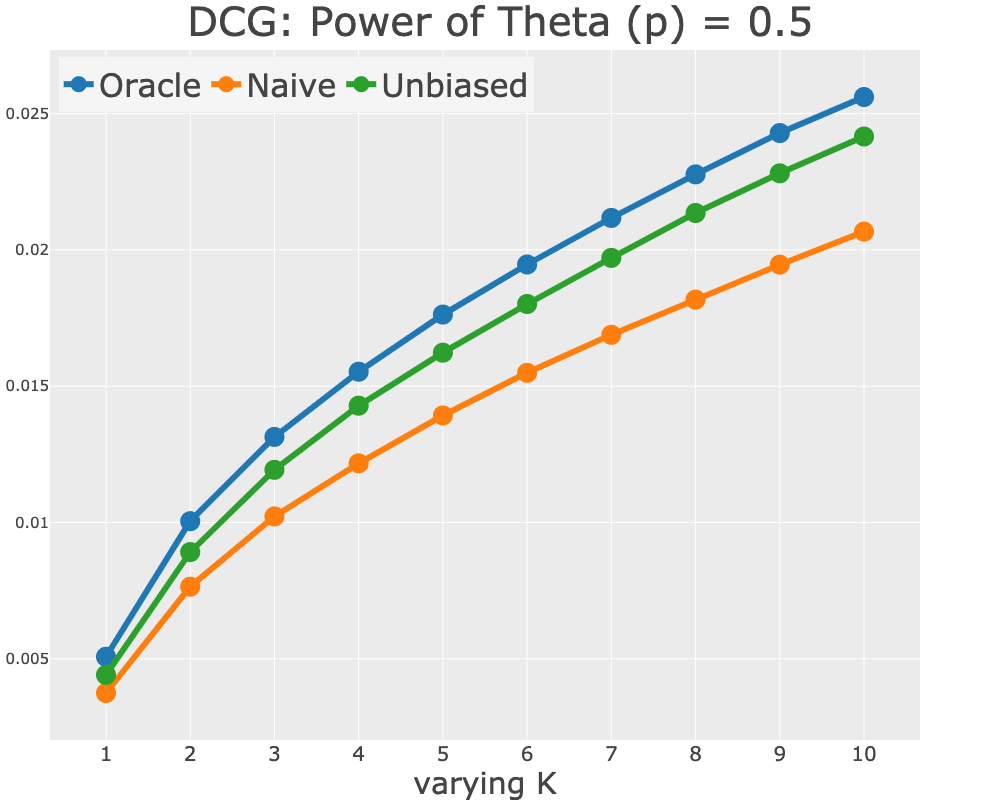}
                \end{center}
            \end{minipage}
            
            \begin{minipage}{0.25\hsize}
                \begin{center}
                    \includegraphics[clip, width=4.5cm]{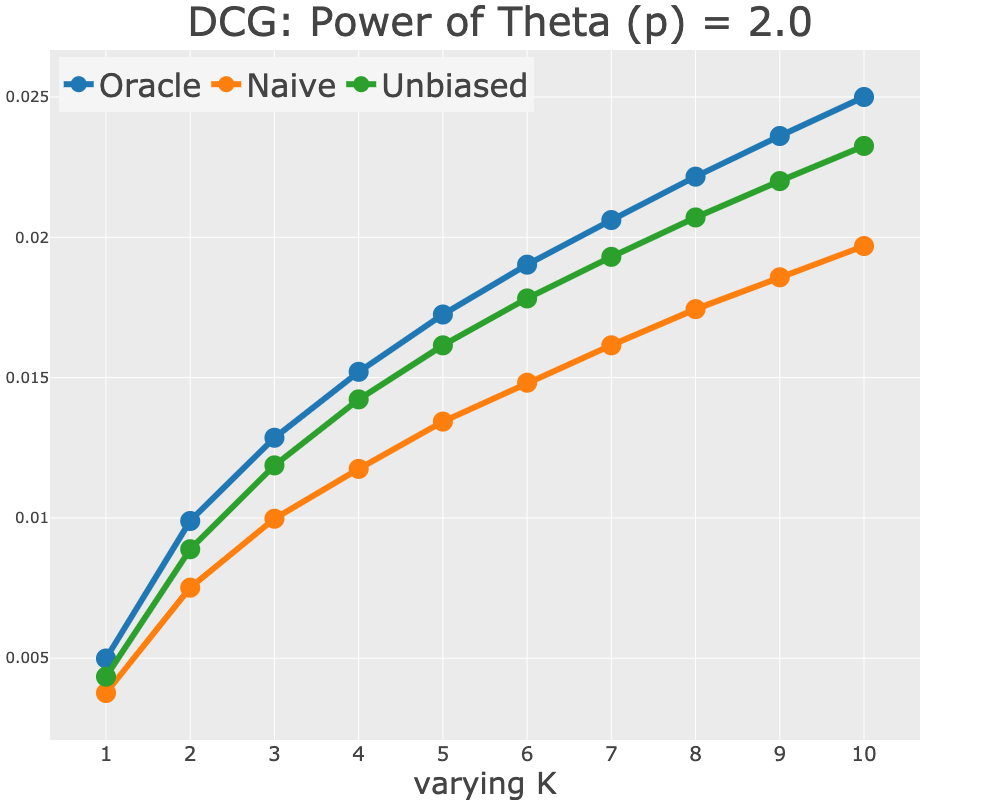}
                \end{center}
            \end{minipage}
            
            \begin{minipage}{0.25\hsize}
                \begin{center}
                    \includegraphics[clip, width=4.5cm]{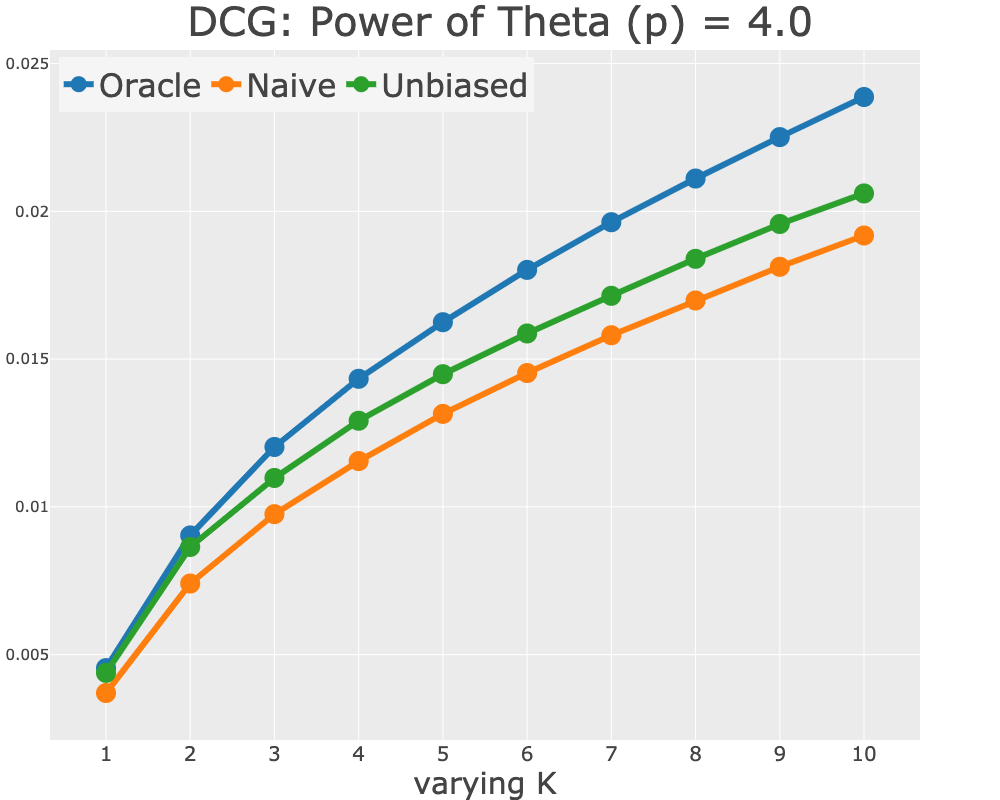}
                \end{center}
            \end{minipage} 
        \end{tabular}
    \end{center}
    \caption{(Left) Log-loss on test sets of the MF-Naive and MF-Unbiased models relative to that of the MF-Oracle with different values of $p$ (x-axis). (Others) DCG@K on test sets of the MF-Oracle, MF-Naive, and MF-Unbiased models with different levels of exposure bias.}
\end{figure*}

\subsubsection{Baselines}
We used and compared the following models.
\begin{itemize}
    \item MF-Oracle: The matrix factorization model trained using the ground truth relevance information. Thus, the performance of this method is the best achievable performance.
    \item MF-Naive: The matrix factorization model trained using only the observable click information. As shown in Proposition 3.1, this estimator has a bias, and this bias problem can be severe, particularly when a large value of $p$ is used. 
    \item MF-Unbiased: The matrix factorization model trained with the unbiased loss function defined in \Eqref{eq9}. The variance of the unbiased loss function might be large when a large value of $p$ is used. 
\end{itemize}

\subsubsection{Evaluation Metrics}
We used the log-loss and discounted cumulative gain (DCG) on test sets to evaluate the relevance prediction and the raking performance, respectively. Note that both evaluation metrics are calculated using the true relevance information, which is inaccessible when using a real-world dataset.

\subsection{Results \& Discussions}

\subsubsection{RQ1. How does the level of exposure bias affect the performance of the MF-Naive and MF-Unbiased models?}
Here, we investigate the performance of the MF-Naive and MF-Unbiased models with different levels of exposure bias (different values of $p$). Figure 1 (left) shows the log-loss on test sets of the MF-Naive and MF-Unbiased models relative to that of the MF-Oracle model. We report the results for varying values of $p$ (x-axis).

The figure reveals that the performance of both the MF-Naive and MF-Unbiased models declines with larger values of $p$. The poor performance of the MF-Naive model with severe exposure bias is due to the bias problem (stated in Proposition 3.1). On the other hand, the poor performance of the MF-Unbiased model arises from the variance problem of the propensity weighting technique (stated in Theorem 4.4). Although the benefit of using the unbiased estimator is relatively small when a severe exposure bias exists, it consistently outperforms the naive estimator in all settings. The results demonstrate the effectiveness of the unbiased estimation approach to the different levels of exposure bias.

\subsubsection{RQ2. Does the optimization of the ideal pointwise loss actually lead to a better value of the ranking metric?}
Next, we compared the ranking performance of the MF with that of the different estimators. Figure 1 shows the DCG@K with different values of $K$ ($K \in \{1, 2, \ldots, 10\}$). We report the results with $p=0.5, 2.0, 4$. As shown in the figure, MF-Unbiased outperforms MF-Naive in all settings. In particular, with $p=0.5$, the MF-Unbiased model largely outperforms the MF-Naive model and its performance is close to that of the MF-Oracle model. However, when a large exposure bias exists ($p=4$), the benefit of using the unbiased estimator is relatively small. These results are well correlated with those for the relevance level prediction task, as discussed in Section 5.2.1. Therefore, the results validate that optimizing the ideal pointwise loss is a valid approach to improving recommendation performance with respect to the ranking metrics.

\section{Real-World Experiment}
In this section, we compare the prediction methods based on the proposed estimator and the existing baselines by using a standard real-world dataset\footnote{The code is available at https://github.com/usaito/unbiased-implicit-rec-real.}. In particular, we investigate the following research question.

\begin{itemize}
    \item [RQ3.] How does the proposed unbiased estimator perform compared with other existing methods?
\end{itemize}

\subsection{Experimental Setup}

\subsubsection{Dataset}
We used the Yahoo! R3 dataset\footnote{http://webscope.sandbox.yahoo.com/}. This is an explicit feedback dataset collected from a song recommendation service. As described in \cite{yang2018unbiased}, it contains users' ratings for randomly selected sets of music as a test set, and thus it can be used to measure the recommenders' true performances. The dataset includes explicit feedback data; we treated items rated greater than or equal to 4 as relevant, and the others were considered irrelevant. We used this dataset because it contains the training and test sets with different item and user distributions. Moreover, it contains the explicit feedback data, thus allowing the utilization of the ground truth relevance information in the test set. Therefore, this dataset is appropriate for the evaluation of recommender in the situation where both positive-unlabeled and MNAR problems exist. To the best of our knowledge, this dataset is the only one that satisfies these properties.

\begin{figure*}[t]
    \centering
    \begin{center}
        \begin{tabular}{c}
            \begin{minipage}{0.32\hsize}
                \begin{center}
                    \includegraphics[clip, width=5.7cm]{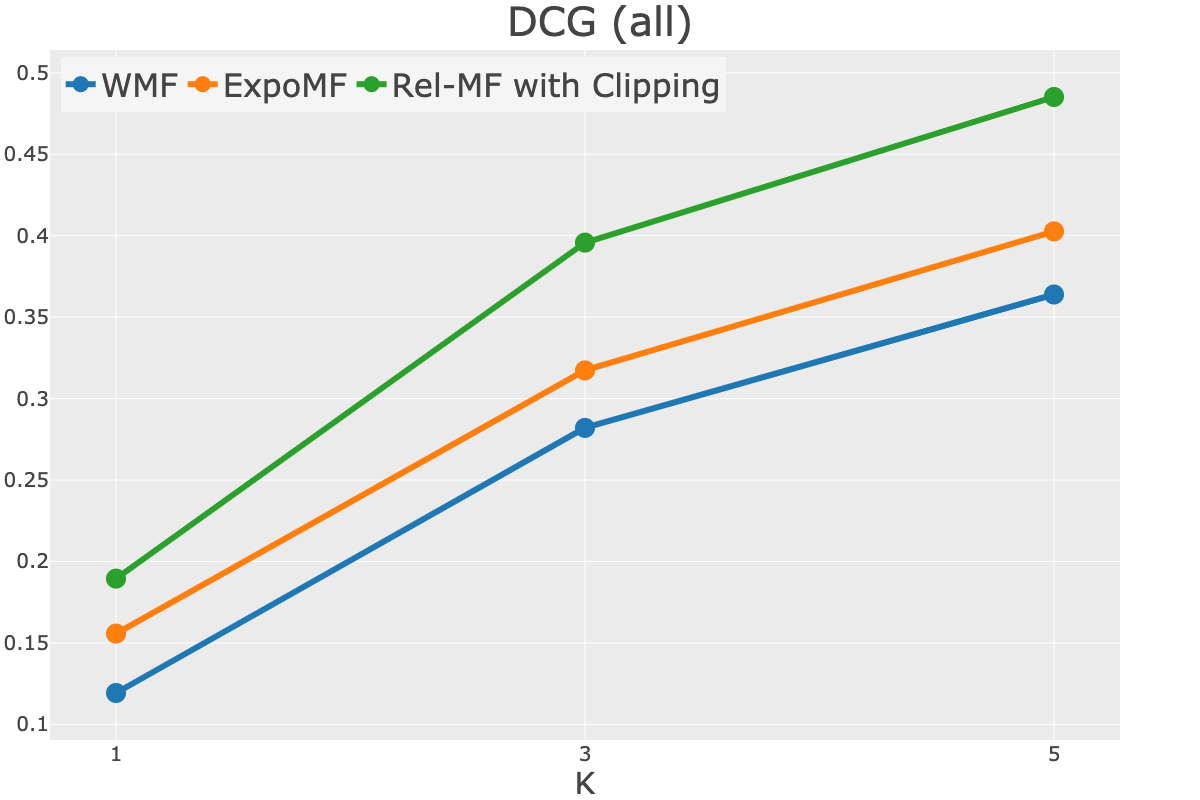}
                \end{center}
            \end{minipage}
            
            \begin{minipage}{0.32\hsize}
                \begin{center}
                    \includegraphics[clip, width=5.7cm]{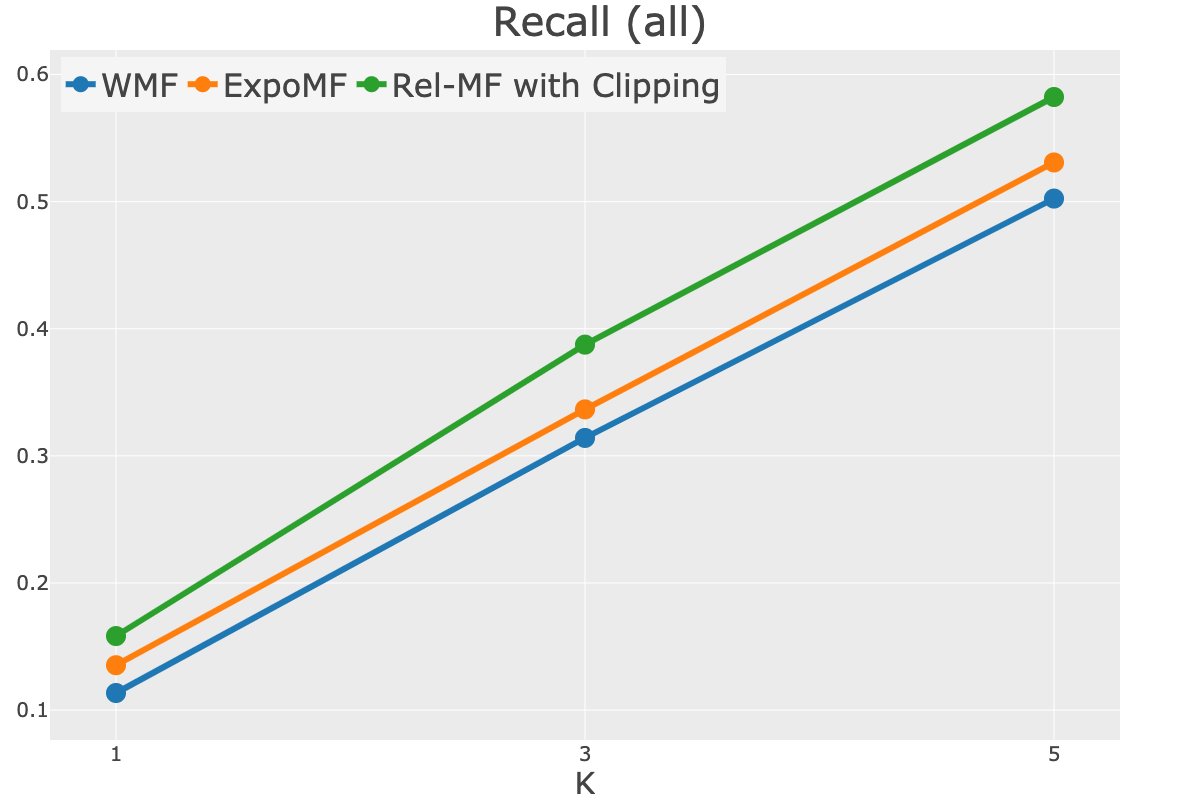}
                \end{center}
            \end{minipage}
            
            \begin{minipage}{0.32\hsize}
                \begin{center}
                    \includegraphics[clip, width=5.7cm]{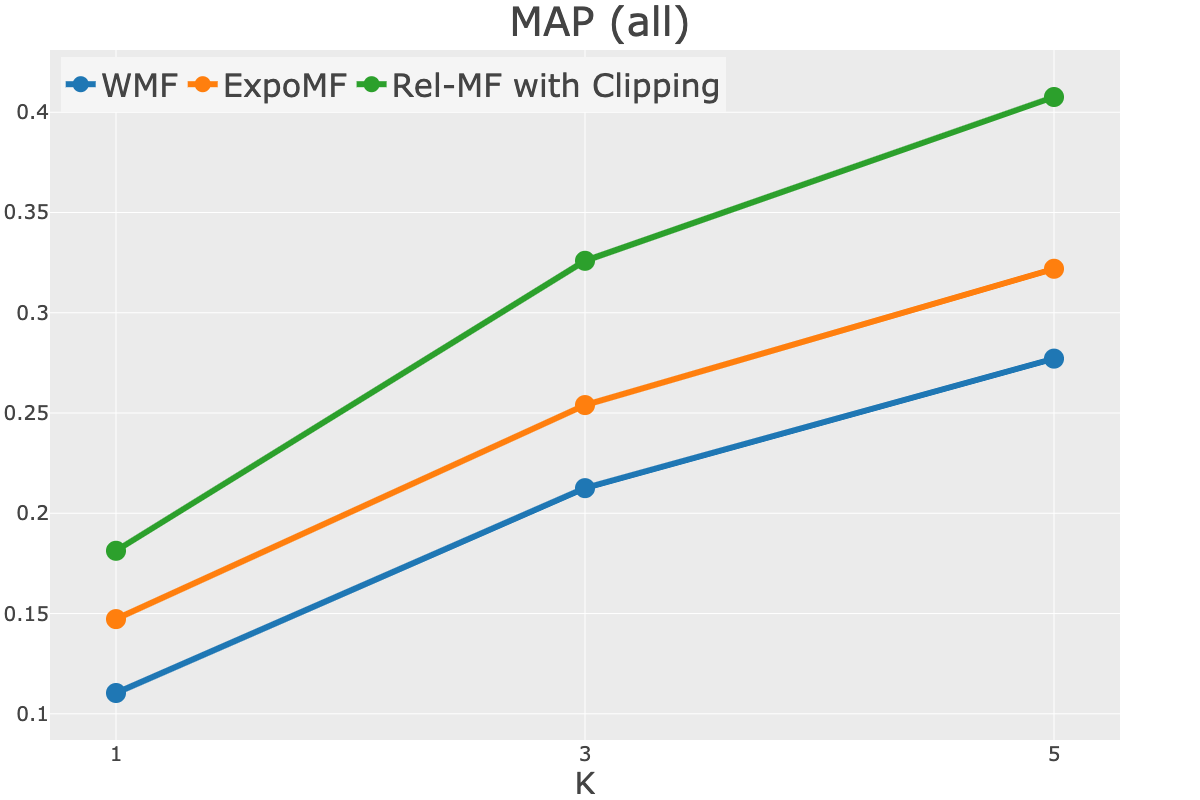}
                \end{center}
            \end{minipage} 
        \end{tabular}
    \end{center}
    
    \centering
    \begin{center}
        \begin{tabular}{c}
            \begin{minipage}{0.32\hsize}
                \begin{center}
                    \includegraphics[clip, width=5.7cm]{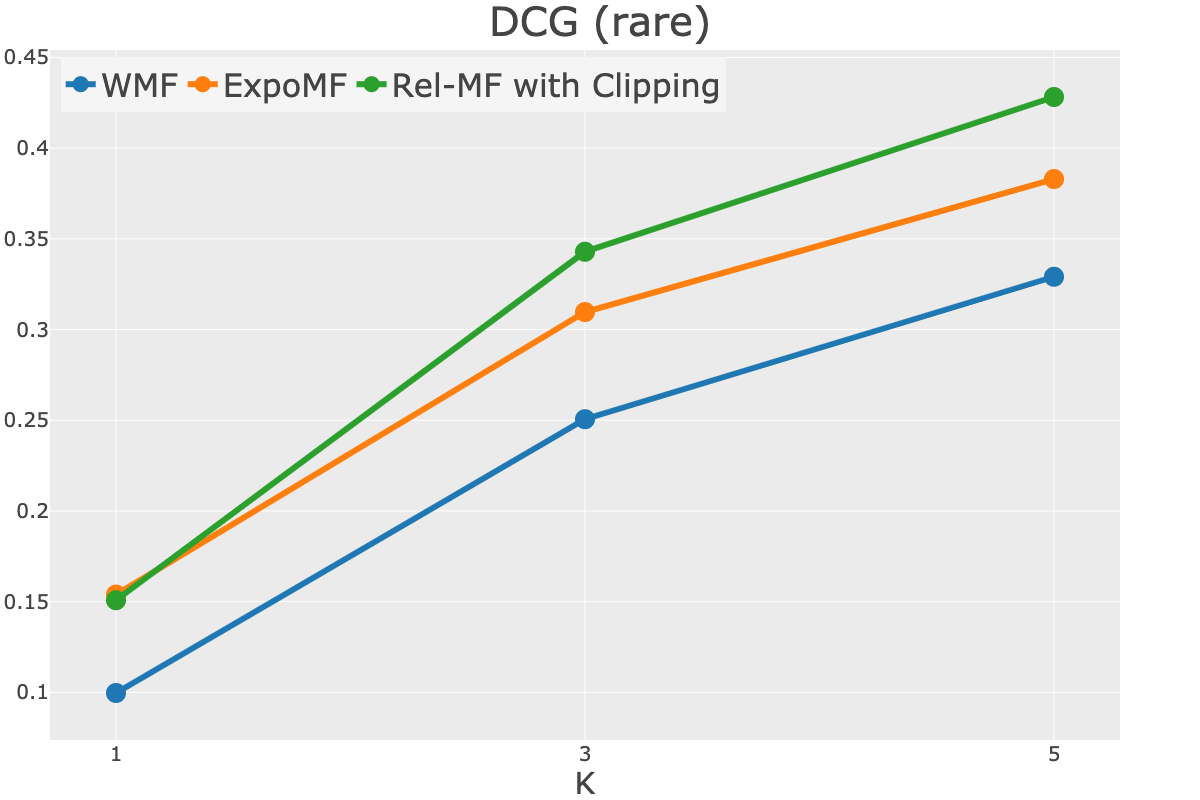}
                \end{center}
            \end{minipage}
            
            \begin{minipage}{0.32\hsize}
                \begin{center}
                    \includegraphics[clip, width=5.7cm]{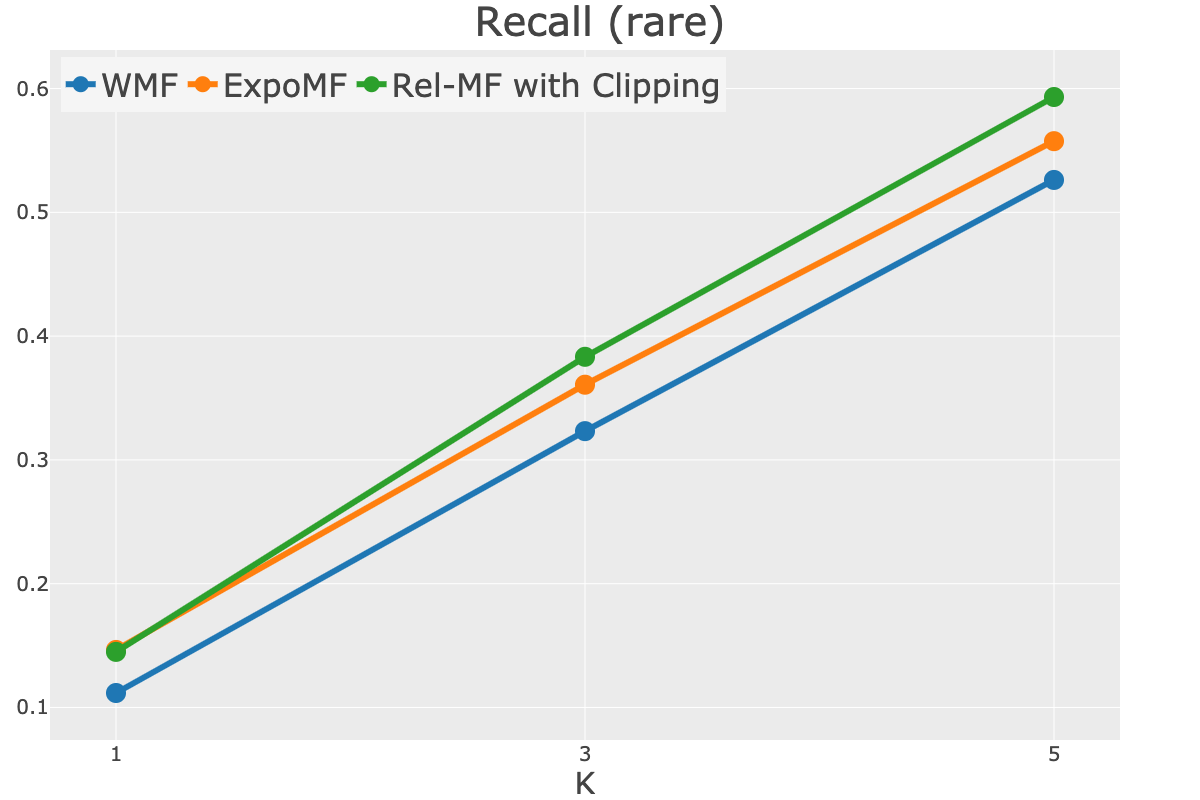}
                \end{center}
            \end{minipage}
            
            \begin{minipage}{0.32\hsize}
                \begin{center}
                    \includegraphics[clip, width=5.7cm]{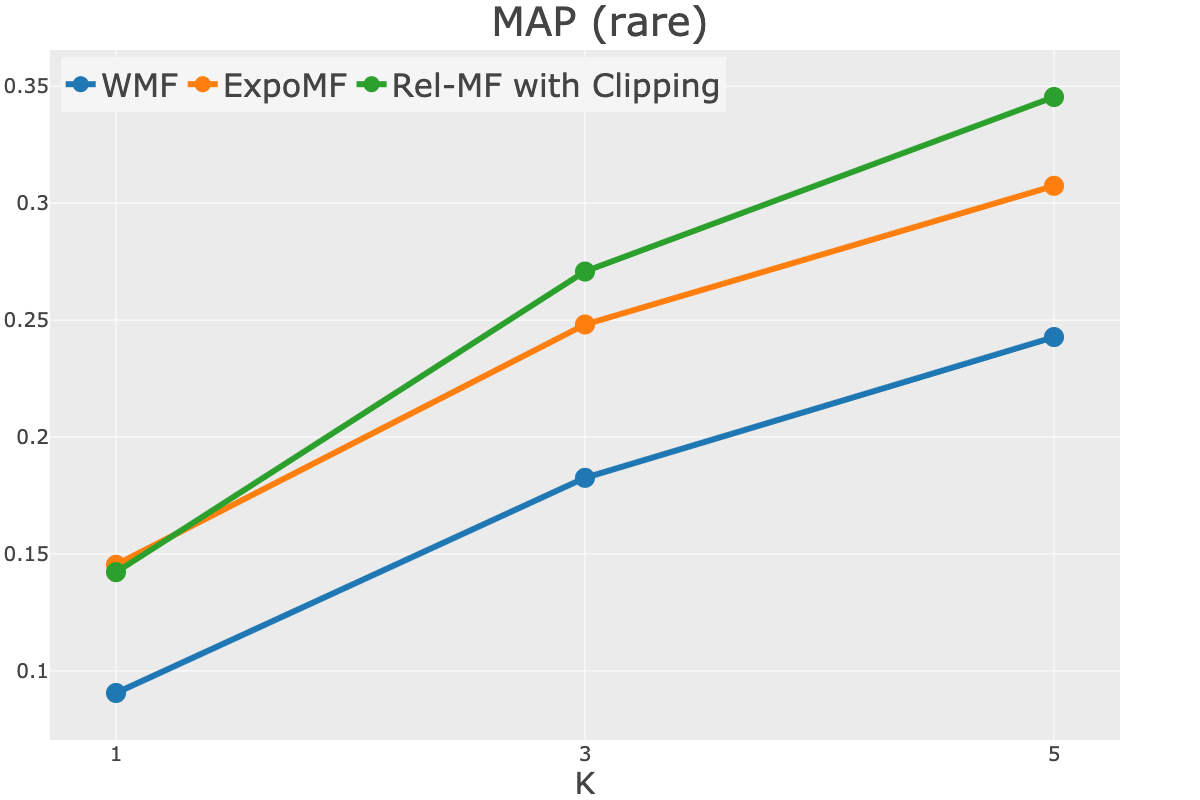}
                \end{center}
            \end{minipage} 
        \end{tabular}
    \end{center}
    
    \caption{Ranking metrics on the Yahoo! R3 data. Rel-MF consistently outperforms the other baselines in almost all situations.}
\end{figure*}

\subsubsection{Baselines and the proposed model}
We describe the existing baselines and the proposed model used in the real-world experiment.

\begin{itemize}
    \item Weighted Matrix Factorization (WMF) \cite{hu2008collaborative}: The WMF is a basic baseline model for implicit recommendation and is described in Section 3.1.
    \item Exposure Matrix Factorization (ExpoMF) \cite{liang2016modeling}: ExpoMF is based on the latent probabilistic model using exposure variables, and it is described in Section 3.2. We used the implementation provided at https://github.com/dawenl/expo-mf.
    \item Relevance Matrix Factorization (Rel-MF): Rel-MF is the proposed model and is based on the same latent factor model as the MF. It updates its user--item factors by minimizing the clipped version of the proposed estimator in \Eqref{eq11}. We estimated the propensity score by using the following relative item popularity. 
    \begin{align}
        \widehat{\theta}_{*, i}=\left(\frac{\sum_{u \in \mathcal{U}} Y_{u, i}}{\max_{i \in \mathcal{I}} \sum_{u \in \mathcal{U}} Y_{u, i}}\right)^{\eta} \label{eq12}
    \end{align}
    In our assumption, the click probability depends on both the exposure probability and relevance level. Thus, we estimated the propensity score using the relative click probability with a parameter $\eta \le 1$ and we set $\eta =0.5$ in this experiment.
\end{itemize}

\subsubsection{Evaluation Metrics} 
We used the DCG, Recall, and MAP to evaluate the ranking performance. We report the results with varying values of $K \in \{ 1, 3, 5 \}$ for all the metrics. These ranking metrics are defined as follows.
\begin{align*}
    DCG@K & = \frac{1}{|\mathcal{U}|} \sum_{u \in \mathcal{U}} \,  \sum_{i \in \mathcal{I}^{test}_u: R_{u,i} = 1} \, \frac{\mathbb{I} \left\{\widehat{Z}_{u,i} \le K \right\} }{\log (\widehat{Z}_{u,i} + 1)}  \\ 
    Recall@K & = \frac{1}{|\mathcal{U}|} \sum_{u \in \mathcal{U}}  \, \sum_{i \in \mathcal{I}^{test}_u: R_{u,i} = 1} \,  \frac{\mathbb{I} \left\{\widehat{Z}_{u,i} \le K \right\} }{\sum_{i \in \mathcal{I}^{test}_u} R_{u,i}}  \\ 
    MAP@K & = \frac{1}{|\mathcal{U}|} \sum_{u \in \mathcal{U}} \, \sum_{i \in \mathcal{I}^{test}_u: R_{u,i} = 1} \, \sum_{k = 1}^K \, \frac{\mathbb{I} \left\{\widehat{Z}_{u,i} \le k \right\} }{k}  
\end{align*}
where $\mathcal{I}^{test}_u$ is a set of items rated by user $u$ in the test set and $R_{u,i}$ takes $1$ if the user rated greater than or equal to $4$.

\subsubsection{Train/Test splitting and hyperparameter tuning criteria}
The original Yahoo! R3 dataset is divided into the training and test sets. We randomly sampled 10\% of positive feedback in the training set as the validation set. Using the validation set, we tuned the dimensions of user--item latent factors within the range of $\{30, 40, \ldots, 200\}$. For the WMF, we tuned the weighting constant in its loss function within the range of $[1, 100]$. In addition, for the Rel-MF with clipping, the clipping hyperparameter $M$ in \Eqref{eq11} was tuned within the range of $[0.01, 0.1]$. For all the baselines and the proposed method, the best set of the hyperparameters was determined using the {\em Optuna} software \cite{akiba2019optuna} with a TPE sampler. For WMF and Rel-MF, we used the squared-loss as $\delta^{(1)}$ and $\delta^{(0)}$ because it is used in the implementation of ExpoMF. Note that the item distributions between the validation and test data are different. Thus, we used the self-normalized inverse propensity score (SNIPS) estimator \cite{yang2018unbiased} of the DCG@5 to tune the hyperparameters. The SNIPS estimator is defined as follows.
\begin{align}
    \widehat{\mathcal{R}}_{SNIPS} \left( \widehat{Z}_{u,i} \right) = \frac{1}{|\mathcal{U}|} \sum_{u \in \mathcal{U}} \, \frac{1}{\sum_{i \in \mathcal{I}_u^{val}} \frac{Y_{u,i}}{\widehat{\theta}_{*, i}}} \sum_{i \in \mathcal{I}_u^{val}} \, \frac{Y_{u,i}}{ \widehat{\theta}_{*, i}} \cdot c \left( \widehat{Z}_{u,i} \right) \notag
\end{align}
where $\mathcal{I}_u^{val}$ is a set of items in the validation set for user $u$\footnote{We randomly sampled 100 unlabeled items for each user to construct $\mathcal{I}_u^{val}$.}. We used the relative item popularity in \Eqref{eq12} as the propensity score of SNIPS estimator.

\subsection{Results \& Discussions}
Here, we present the experimental results in detail.

\subsubsection{RQ3. How does the proposed unbiased estimator perform compared with other existing methods?}

Figure 2 shows the performance of the methods, corresponding to rare and all items. We treated 500 items (out of 1,000) with small relative item popularity in \Eqref{eq12} in the training set as rare. As shown in the figure, the Rel-MF significantly outperformed the other baselines on almost all the metrics, for example, improving the DCG@5 by 20.4\%, Recall@5 by 9.6\%, and MAP@5 by 26.6\% over the ExpoMF. Furthermore, for rare items, the proposed method improved the DCG@5 by 11.8\%, Recall@5 by 6.4\%, and MAP@5 by 11.2\% over the best baseline. However, for all metrics on rare items, the Rel-MF was outperformed by ExpoMF for $K=1$. The propensity estimation might be the reason for this. We estimated the user independent propensity score by \Eqref{eq12}. However, in real-world recommender systems, the user activeness can be diverse, and different users can have different propensity scores for the same item. Thus, considering only the item dependent propensity score might be too simple.

Overall, it was observed that the ExpoMF outperformed the WMF in all settings, which is consistent with the previous experiments \cite{liang2016modeling,2018DynamicMissingness}. Moreover, Rel-MF outperformed the other baseline methods in most cases. In addition, the proposed method improved the ranking metrics for less-frequently observed items in the training sets (rare items). This is because the ExpoMF downweights the prediction losses on the items having a low exposure probability; in contrast, the proposed method utilizes the theoretically principal estimation technique and solves the MNAR problem by ensuring prediction accuracy on rare items. These results suggest that the proposed method is the most suitable method for optimizing the metric of interest defined in \Eqref{eq4} from biased implicit feedback.

\section{Conclusion}
In this study, we first defined the ideal loss function for maximizing the relevance to optimize the user experience. Subsequently, we demonstrated that the loss functions of WMF and ExpoMF are biased toward the ideal loss. Furthermore, we proposed an unbiased estimator for the ideal loss inspired by the estimation method for causal inference and positive-unlabeled learning. We also analyzed the variance of the unbiased estimator and introduced a clipped estimator, which, by introducing a small bias, could reduce the variance and achieve better performance as a result of a better bias-variance trade-off. In the experiments, the proposed method significantly outperformed the existing methods with respect to the relevance maximization. In particular, the proposed method outperformed these methods for items with low exposure probability, and this finding empirically suggests that the proposed approach can suitably maximize the user experience.

A possible next step would be the development of a sophisticated method to estimate the exposure probabilities. The proposed unbiased estimator relies on the true propensity scores for its unbiasedness. Thus, a better estimation of the exposure probabilities could lead to better prediction performances owing to the improved estimation of the loss of interest. Moreover, pairwise algorithms that address the two challenging problems of implicit feedback have not yet been proposed. Therefore, the extension of the unbiased estimator to the pairwise algorithm is another interesting theme.

\bibliographystyle{ACM-Reference-Format}
\input{relmf.bbl}

\end{document}

%% file: relmf.bbl

%% file: main.bbl
\begin{thebibliography}{28}

%%% the Web address.

\ifx \showCODEN    \undefined \def \showCODEN     #1{\unskip}     \fi
\ifx \showDOI      \undefined \def \showDOI       #1{#1}\fi
\ifx \showISBNx    \undefined \def \showISBNx     #1{\unskip}     \fi
\ifx \showISBNxiii \undefined \def \showISBNxiii  #1{\unskip}     \fi
\ifx \showISSN     \undefined \def \showISSN      #1{\unskip}     \fi
\ifx \showLCCN     \undefined \def \showLCCN      #1{\unskip}     \fi
\ifx \shownote     \undefined \def \shownote      #1{#1}          \fi
\ifx \showarticletitle \undefined \def \showarticletitle #1{#1}   \fi
\ifx \showURL      \undefined \def \showURL       {\relax}        \fi
% invisible to TeX
\providecommand\bibfield[2]{#2}
\providecommand\bibinfo[2]{#2}
\providecommand\natexlab[1]{#1}
\providecommand\showeprint[2][]{arXiv:#2}

\bibitem[\protect\citeauthoryear{Akiba, Sano, Yanase, Ohta, and Koyama}{Akiba
  et~al\mbox{.}}{2019}]%
        {akiba2019optuna}
\bibfield{author}{\bibinfo{person}{Takuya Akiba}, \bibinfo{person}{Shotaro
  Sano}, \bibinfo{person}{Toshihiko Yanase}, \bibinfo{person}{Takeru Ohta},
  {and} \bibinfo{person}{Masanori Koyama}.} \bibinfo{year}{2019}\natexlab{}.
\newblock \showarticletitle{Optuna: A Next-generation Hyperparameter
  Optimization Framework}. In \bibinfo{booktitle}{\emph{Proceedings of the 25th
  ACM SIGKDD International Conference on Knowledge Discovery \& Data Mining}}.
  ACM, \bibinfo{pages}{2623--2631}.
\newblock


\bibitem[\protect\citeauthoryear{Bekker and Davis}{Bekker and Davis}{2018}]%
        {bekker2018beyond}
\bibfield{author}{\bibinfo{person}{Jessa Bekker} {and} \bibinfo{person}{Jesse
  Davis}.} \bibinfo{year}{2018}\natexlab{}.
\newblock \showarticletitle{Beyond the selected completely at random assumption
  for learning from positive and unlabeled data}.
\newblock \bibinfo{journal}{\emph{arXiv preprint arXiv:1809.03207}}
  (\bibinfo{year}{2018}).
\newblock


\bibitem[\protect\citeauthoryear{Bonner and Vasile}{Bonner and Vasile}{2018}]%
        {bonner2018causal}
\bibfield{author}{\bibinfo{person}{Stephen Bonner} {and}
  \bibinfo{person}{Flavian Vasile}.} \bibinfo{year}{2018}\natexlab{}.
\newblock \showarticletitle{Causal Embeddings for Recommendation}. In
  \bibinfo{booktitle}{\emph{Proceedings of the 12th ACM Conference on
  Recommender Systems}} \emph{(\bibinfo{series}{RecSys '18})}.
  \bibinfo{publisher}{ACM}, \bibinfo{address}{New York, NY, USA},
  \bibinfo{pages}{104--112}.
\newblock
\showISBNx{978-1-4503-5901-6}
\urldef\tempurl%
\url{https://doi.org/10.1145/3240323.3240360}
\showDOI{\tempurl}


\bibitem[\protect\citeauthoryear{Chen, Beutel, Covington, Jain, Belletti, and
  Chi}{Chen et~al\mbox{.}}{2019}]%
        {chen2019top}
\bibfield{author}{\bibinfo{person}{Minmin Chen}, \bibinfo{person}{Alex Beutel},
  \bibinfo{person}{Paul Covington}, \bibinfo{person}{Sagar Jain},
  \bibinfo{person}{Francois Belletti}, {and} \bibinfo{person}{Ed~H Chi}.}
  \bibinfo{year}{2019}\natexlab{}.
\newblock \showarticletitle{Top-k off-policy correction for a REINFORCE
  recommender system}. In \bibinfo{booktitle}{\emph{Proceedings of the Twelfth
  ACM International Conference on Web Search and Data Mining}}. ACM,
  \bibinfo{pages}{456--464}.
\newblock


\bibitem[\protect\citeauthoryear{Elkan and Noto}{Elkan and Noto}{2008}]%
        {elkan2008learning}
\bibfield{author}{\bibinfo{person}{Charles Elkan} {and} \bibinfo{person}{Keith
  Noto}.} \bibinfo{year}{2008}\natexlab{}.
\newblock \showarticletitle{Learning classifiers from only positive and
  unlabeled data}. In \bibinfo{booktitle}{\emph{Proceedings of the 14th ACM
  SIGKDD international conference on Knowledge discovery and data mining}}.
  ACM, \bibinfo{pages}{213--220}.
\newblock


\bibitem[\protect\citeauthoryear{Gilotte, Calauz{\`e}nes, Nedelec, Abraham, and
  Doll{\'e}}{Gilotte et~al\mbox{.}}{2018}]%
        {gilotte2018offline}
\bibfield{author}{\bibinfo{person}{Alexandre Gilotte},
  \bibinfo{person}{Cl{\'e}ment Calauz{\`e}nes}, \bibinfo{person}{Thomas
  Nedelec}, \bibinfo{person}{Alexandre Abraham}, {and} \bibinfo{person}{Simon
  Doll{\'e}}.} \bibinfo{year}{2018}\natexlab{}.
\newblock \showarticletitle{Offline a/b testing for recommender systems}. In
  \bibinfo{booktitle}{\emph{Proceedings of the Eleventh ACM International
  Conference on Web Search and Data Mining}}. ACM, \bibinfo{pages}{198--206}.
\newblock


\bibitem[\protect\citeauthoryear{Hu, Koren, and Volinsky}{Hu
  et~al\mbox{.}}{2008}]%
        {hu2008collaborative}
\bibfield{author}{\bibinfo{person}{Yifan Hu}, \bibinfo{person}{Yehuda Koren},
  {and} \bibinfo{person}{Chris Volinsky}.} \bibinfo{year}{2008}\natexlab{}.
\newblock \showarticletitle{Collaborative filtering for implicit feedback
  datasets}. In \bibinfo{booktitle}{\emph{2008 Eighth IEEE International
  Conference on Data Mining}}. Ieee, \bibinfo{pages}{263--272}.
\newblock


\bibitem[\protect\citeauthoryear{Imbens and Rubin}{Imbens and Rubin}{2015}]%
        {imbens2015causal}
\bibfield{author}{\bibinfo{person}{Guido~W Imbens} {and}
  \bibinfo{person}{Donald~B Rubin}.} \bibinfo{year}{2015}\natexlab{}.
\newblock \bibinfo{booktitle}{\emph{Causal inference in statistics, social, and
  biomedical sciences}}.
\newblock \bibinfo{publisher}{Cambridge University Press}.
\newblock


\bibitem[\protect\citeauthoryear{Jannach~D.}{Jannach~D.}{2018}]%
        {2018ImplicitReview}
\bibfield{author}{\bibinfo{person}{Zanker~M Jannach~D., Lerche~L.}}
  \bibinfo{year}{2018}\natexlab{}.
\newblock \bibinfo{booktitle}{\emph{Recommending Based on Implicit Feedback}}.
\newblock \bibinfo{publisher}{Springer}.
\newblock


\bibitem[\protect\citeauthoryear{Jiahui~Liu}{Jiahui~Liu}{2010}]%
        {2010LiuClick}
\bibfield{author}{\bibinfo{person}{Elin Rønby~Pedersen Jiahui~Liu,
  Peter~Dolan}.} \bibinfo{year}{2010}\natexlab{}.
\newblock \showarticletitle{Personalized news recommendation based on click
  behavior.}. In \bibinfo{booktitle}{\emph{Proc. of 14th Int. Conf. on
  Intelligent User Interfaces (IUI)}}. ACM, \bibinfo{pages}{31--40}.
\newblock


\bibitem[\protect\citeauthoryear{Joachims and Swaminathan}{Joachims and
  Swaminathan}{2016}]%
        {joachims2016counterfactual}
\bibfield{author}{\bibinfo{person}{Thorsten Joachims} {and}
  \bibinfo{person}{Adith Swaminathan}.} \bibinfo{year}{2016}\natexlab{}.
\newblock \showarticletitle{Counterfactual evaluation and learning for search,
  recommendation and ad placement}. In \bibinfo{booktitle}{\emph{Proceedings of
  the 39th International ACM SIGIR conference on Research and Development in
  Information Retrieval}}. ACM, \bibinfo{pages}{1199--1201}.
\newblock


\bibitem[\protect\citeauthoryear{Joachims, Swaminathan, and Schnabel}{Joachims
  et~al\mbox{.}}{2017}]%
        {joachims2017unbiased}
\bibfield{author}{\bibinfo{person}{Thorsten Joachims}, \bibinfo{person}{Adith
  Swaminathan}, {and} \bibinfo{person}{Tobias Schnabel}.}
  \bibinfo{year}{2017}\natexlab{}.
\newblock \showarticletitle{Unbiased learning-to-rank with biased feedback}. In
  \bibinfo{booktitle}{\emph{Proceedings of the Tenth ACM International
  Conference on Web Search and Data Mining}}. ACM, \bibinfo{pages}{781--789}.
\newblock


\bibitem[\protect\citeauthoryear{Johnson}{Johnson}{2014}]%
        {johnson2014logistic}
\bibfield{author}{\bibinfo{person}{Christopher~C Johnson}.}
  \bibinfo{year}{2014}\natexlab{}.
\newblock \showarticletitle{Logistic matrix factorization for implicit feedback
  data}.
\newblock \bibinfo{journal}{\emph{Advances in Neural Information Processing
  Systems}}  \bibinfo{volume}{27} (\bibinfo{year}{2014}).
\newblock


\bibitem[\protect\citeauthoryear{Koren, Bell, and Volinsky}{Koren
  et~al\mbox{.}}{2009}]%
        {koren2009matrix}
\bibfield{author}{\bibinfo{person}{Yehuda Koren}, \bibinfo{person}{Robert
  Bell}, {and} \bibinfo{person}{Chris Volinsky}.}
  \bibinfo{year}{2009}\natexlab{}.
\newblock \showarticletitle{Matrix factorization techniques for recommender
  systems}.
\newblock \bibinfo{journal}{\emph{Computer}} \bibinfo{number}{8}
  (\bibinfo{year}{2009}), \bibinfo{pages}{30--37}.
\newblock


\bibitem[\protect\citeauthoryear{Li and Liu}{Li and Liu}{2005}]%
        {li2005learning}
\bibfield{author}{\bibinfo{person}{Xiao-Li Li} {and} \bibinfo{person}{Bing
  Liu}.} \bibinfo{year}{2005}\natexlab{}.
\newblock \showarticletitle{Learning from positive and unlabeled examples with
  different data distributions}. In \bibinfo{booktitle}{\emph{European
  Conference on Machine Learning}}. Springer, \bibinfo{pages}{218--229}.
\newblock


\bibitem[\protect\citeauthoryear{Liang, Altosaar, Charlin, and Blei}{Liang
  et~al\mbox{.}}{2016a}]%
        {liang2016factorization}
\bibfield{author}{\bibinfo{person}{Dawen Liang}, \bibinfo{person}{Jaan
  Altosaar}, \bibinfo{person}{Laurent Charlin}, {and} \bibinfo{person}{David~M
  Blei}.} \bibinfo{year}{2016}\natexlab{a}.
\newblock \showarticletitle{Factorization meets the item embedding:
  Regularizing matrix factorization with item co-occurrence}. In
  \bibinfo{booktitle}{\emph{Proceedings of the 10th ACM conference on
  recommender systems}}. ACM, \bibinfo{pages}{59--66}.
\newblock


\bibitem[\protect\citeauthoryear{Liang, Charlin, McInerney, and Blei}{Liang
  et~al\mbox{.}}{2016b}]%
        {liang2016modeling}
\bibfield{author}{\bibinfo{person}{Dawen Liang}, \bibinfo{person}{Laurent
  Charlin}, \bibinfo{person}{James McInerney}, {and} \bibinfo{person}{David~M
  Blei}.} \bibinfo{year}{2016}\natexlab{b}.
\newblock \showarticletitle{Modeling user exposure in recommendation}. In
  \bibinfo{booktitle}{\emph{Proceedings of the 25th International Conference on
  World Wide Web}}. International World Wide Web Conferences Steering
  Committee, \bibinfo{pages}{951--961}.
\newblock


\bibitem[\protect\citeauthoryear{Liu, Lin, Zhang, Xiao, and Tong}{Liu
  et~al\mbox{.}}{2019}]%
        {liu2019spiral}
\bibfield{author}{\bibinfo{person}{Dugang Liu}, \bibinfo{person}{Chen Lin},
  \bibinfo{person}{Zhilin Zhang}, \bibinfo{person}{Yanghua Xiao}, {and}
  \bibinfo{person}{Hanghang Tong}.} \bibinfo{year}{2019}\natexlab{}.
\newblock \showarticletitle{Spiral of Silence in Recommender Systems}. In
  \bibinfo{booktitle}{\emph{Proceedings of the Twelfth ACM International
  Conference on Web Search and Data Mining}}. ACM, \bibinfo{pages}{222--230}.
\newblock


\bibitem[\protect\citeauthoryear{Menghan~Wang and Zhang}{Menghan~Wang and
  Zhang}{2018}]%
        {2018Wangsocialexposure:}
\bibfield{author}{\bibinfo{person}{Yang~Yang Menghan~Wang, Xiaolin~Zheng} {and}
  \bibinfo{person}{Kun Zhang}.} \bibinfo{year}{2018}\natexlab{}.
\newblock \showarticletitle{Collaborative filtering with social exposure: A
  modular approach to social recommendation}. In \bibinfo{booktitle}{\emph{The
  Thirty-Second AAAI Conference on Artificial Intelligence}}. AAAI,
  \bibinfo{pages}{2516--2523}.
\newblock


\bibitem[\protect\citeauthoryear{Rosenbaum and Rubin}{Rosenbaum and
  Rubin}{1983}]%
        {rosenbaum1983central}
\bibfield{author}{\bibinfo{person}{Paul~R Rosenbaum} {and}
  \bibinfo{person}{Donald~B Rubin}.} \bibinfo{year}{1983}\natexlab{}.
\newblock \showarticletitle{The central role of the propensity score in
  observational studies for causal effects}.
\newblock \bibinfo{journal}{\emph{Biometrika}} \bibinfo{volume}{70},
  \bibinfo{number}{1} (\bibinfo{year}{1983}), \bibinfo{pages}{41--55}.
\newblock


\bibitem[\protect\citeauthoryear{Rubin}{Rubin}{1974}]%
        {rubin1974estimating}
\bibfield{author}{\bibinfo{person}{Donald~B Rubin}.}
  \bibinfo{year}{1974}\natexlab{}.
\newblock \showarticletitle{Estimating causal effects of treatments in
  randomized and nonrandomized studies.}
\newblock \bibinfo{journal}{\emph{Journal of educational Psychology}}
  \bibinfo{volume}{66}, \bibinfo{number}{5} (\bibinfo{year}{1974}),
  \bibinfo{pages}{688}.
\newblock


\bibitem[\protect\citeauthoryear{Saito, Sakata, and Nakata}{Saito
  et~al\mbox{.}}{2019}]%
        {saito2019doubly}
\bibfield{author}{\bibinfo{person}{Yuta Saito}, \bibinfo{person}{Hayato
  Sakata}, {and} \bibinfo{person}{Kazuhide Nakata}.}
  \bibinfo{year}{2019}\natexlab{}.
\newblock \showarticletitle{Doubly Robust Prediction and Evaluation Methods
  Improve Uplift Modeling for Observational Data}. In
  \bibinfo{booktitle}{\emph{Proceedings of the 2019 SIAM International
  Conference on Data Mining}}. SIAM, \bibinfo{pages}{468--476}.
\newblock


\bibitem[\protect\citeauthoryear{Schnabel, Swaminathan, Singh, Chandak, and
  Joachims}{Schnabel et~al\mbox{.}}{2016}]%
        {schnabel2016recommendations}
\bibfield{author}{\bibinfo{person}{Tobias Schnabel}, \bibinfo{person}{Adith
  Swaminathan}, \bibinfo{person}{Ashudeep Singh}, \bibinfo{person}{Navin
  Chandak}, {and} \bibinfo{person}{Thorsten Joachims}.}
  \bibinfo{year}{2016}\natexlab{}.
\newblock \showarticletitle{Recommendations as Treatments: Debiasing Learning
  and Evaluation}. In \bibinfo{booktitle}{\emph{Proceedings of The 33rd
  International Conference on Machine Learning}}
  \emph{(\bibinfo{series}{Proceedings of Machine Learning Research})},
  \bibfield{editor}{\bibinfo{person}{Maria~Florina Balcan} {and}
  \bibinfo{person}{Kilian~Q. Weinberger}} (Eds.), Vol.~\bibinfo{volume}{48}.
  \bibinfo{publisher}{PMLR}, \bibinfo{address}{New York, New York, USA},
  \bibinfo{pages}{1670--1679}.
\newblock
\urldef\tempurl%
\url{http://proceedings.mlr.press/v48/schnabel16.html}
\showURL{%
\tempurl}


\bibitem[\protect\citeauthoryear{Su, Wang, Santacatterina, and Joachims}{Su
  et~al\mbox{.}}{2019}]%
        {su2019cab}
\bibfield{author}{\bibinfo{person}{Yi Su}, \bibinfo{person}{Lequn Wang},
  \bibinfo{person}{Michele Santacatterina}, {and} \bibinfo{person}{Thorsten
  Joachims}.} \bibinfo{year}{2019}\natexlab{}.
\newblock \showarticletitle{CAB: Continuous Adaptive Blending for Policy
  Evaluation and Learning}. In \bibinfo{booktitle}{\emph{International
  Conference on Machine Learning}}. \bibinfo{pages}{6005--6014}.
\newblock


\bibitem[\protect\citeauthoryear{Swaminathan and Joachims}{Swaminathan and
  Joachims}{2015}]%
        {swaminathan2015self}
\bibfield{author}{\bibinfo{person}{Adith Swaminathan} {and}
  \bibinfo{person}{Thorsten Joachims}.} \bibinfo{year}{2015}\natexlab{}.
\newblock \showarticletitle{The self-normalized estimator for counterfactual
  learning}. In \bibinfo{booktitle}{\emph{advances in neural information
  processing systems}}. \bibinfo{pages}{3231--3239}.
\newblock


\bibitem[\protect\citeauthoryear{Wang and Zhang}{Wang and Zhang}{2018}]%
        {2018DynamicMissingness}
\bibfield{author}{\bibinfo{person}{Gong M. Zheng~X. Wang, M.} {and}
  \bibinfo{person}{K Zhang}.} \bibinfo{year}{2018}\natexlab{}.
\newblock \showarticletitle{Modeling dynamic missingness of implicit feedback
  for recommendation.}. In \bibinfo{booktitle}{\emph{Conference on Neural
  Information Processing Systems}}.
\newblock


\bibitem[\protect\citeauthoryear{Wang, Golbandi, Bendersky, Metzler, and
  Najork}{Wang et~al\mbox{.}}{2018}]%
        {wang2018position}
\bibfield{author}{\bibinfo{person}{Xuanhui Wang}, \bibinfo{person}{Nadav
  Golbandi}, \bibinfo{person}{Michael Bendersky}, \bibinfo{person}{Donald
  Metzler}, {and} \bibinfo{person}{Marc Najork}.}
  \bibinfo{year}{2018}\natexlab{}.
\newblock \showarticletitle{Position bias estimation for unbiased learning to
  rank in personal search}. In \bibinfo{booktitle}{\emph{Proceedings of the
  Eleventh ACM International Conference on Web Search and Data Mining}}. ACM,
  \bibinfo{pages}{610--618}.
\newblock


\bibitem[\protect\citeauthoryear{Yang, Cui, Xuan, Wang, Belongie, and
  Estrin}{Yang et~al\mbox{.}}{2018}]%
        {yang2018unbiased}
\bibfield{author}{\bibinfo{person}{Longqi Yang}, \bibinfo{person}{Yin Cui},
  \bibinfo{person}{Yuan Xuan}, \bibinfo{person}{Chenyang Wang},
  \bibinfo{person}{Serge Belongie}, {and} \bibinfo{person}{Deborah Estrin}.}
  \bibinfo{year}{2018}\natexlab{}.
\newblock \showarticletitle{Unbiased Offline Recommender Evaluation for
  Missing-not-at-random Implicit Feedback}. In
  \bibinfo{booktitle}{\emph{Proceedings of the 12th ACM Conference on
  Recommender Systems}} \emph{(\bibinfo{series}{RecSys '18})}.
  \bibinfo{publisher}{ACM}, \bibinfo{address}{New York, NY, USA},
  \bibinfo{pages}{279--287}.
\newblock
\showISBNx{978-1-4503-5901-6}
\urldef\tempurl%
\url{https://doi.org/10.1145/3240323.3240355}
\showDOI{\tempurl}


\end{thebibliography}
